\documentclass[journal,onecolumn,11pt,draftclsnofoot]{IEEEtran}

\def\Figs{./figs/} 

\usepackage{graphicx}
\usepackage{balance}
\usepackage[utf8]{inputenc} 
\usepackage[T1]{fontenc}
\usepackage{url}
\usepackage{ifthen}
\usepackage{comment}
\usepackage[cmex10]{amsmath} 

\newboolean{arXiv} 
\setboolean{arXiv}{true} 

\newboolean{NOTE} 
\setboolean{NOTE}{false} 

\usepackage[url,hyperrefblack,notheorems,IEEEtran]{research17} 
\def\Figs{figs/}

\usepackage{amsthm} 
\newtheorem{theorem}{\mytheoremname}

\newtheorem{proposition}{\mypropositionname}

\newtheorem{definition}{\mydefinitionname}

\usepackage{subcaption} 
\usepackage[capitalize]{cleveref}
\allowdisplaybreaks

\usepackage{tikz} 
\usetikzlibrary{calc, arrows.meta, intersections, patterns, positioning, shapes.misc, fadings, through,decorations.pathreplacing}

\definecolor{MidnightBlue}{HTML}{006795}
\definecolor{RedOrange}{HTML}{F26035}
\definecolor{Plum}{HTML}{92268F}
\definecolor{Dandelion}{HTML}{00CD00}
\newcommand{\Lap}[1]{\tn{Lap}\left(#1\right)} 
\newcommand{\eLap}[1]{\tn{Lap}(#1)}

\newcommand{\dTV}[2]{d_{\textnormal{TV}}\left(#1 \kern0.1em{,}\kern0.1em #2\right)}
\newcommand{\edTV}[2]{d_{\textnormal{TV}}(#1 \kern0.1em{,}\kern0.1em #2)}
\newcommand{\bigdTV}[2]{d_{\textnormal{TV}}\bigl(#1 \kern0.1em{,}\kern0.1em #2\bigr)}

\newcommand{\Loss}{\mathsf{L}}
\newcommand{\avgLoss}{\bar{\mathsf{L}}}
\newcommand{\Risk}{\mathsf{R}}
\newcommand{\txtc}{\textnormal{c}}
\newcommand{\tc}[1]{t_{\textnormal{c},#1}}
\newcommand{\epsc}[1]{\eps_{\textnormal{c},#1}}
\newcommand{\tagg}{t_{\textnormal{agg}}}
\renewcommand{\vect}[1]{#1} 
\usepackage{soul} 
\usepackage[colorinlistoftodos, textsize=footnotesize]{todonotes} 
\definecolor{darkgreen}{rgb}{0.4, 0.7, 0.0}
\newcommand{\hy}{\color{black}} 
\begin{document}
\title{Age Aware Scheduling for Differentially-Private Federated Learning}


\ifthenelse{\boolean{arXiv}}{\author{Kuan-Yu~Lin,~\IEEEmembership{Student Member,~IEEE,}, Hsuan-Yin~Lin,~\IEEEmembership{Senior Member,~IEEE}, Yu-Pin~Hsu~\IEEEmembership{Senior Member,~IEEE}, and Yu-Chih~Huang~\IEEEmembership{Senior Member,~IEEE}
\thanks{This paper was presented in part at the IEEE International Symposium on Information Theory (ISIT), Athens, Greece, July 2024~\cite{LinLinHsuHuang24_1app}.}
\thanks{\hy K.-Y.~Lin and Y.-C.~Huang are with the Institute of Communications Engineering, National Yang Ming Chiao Tung University, Taiwan (email: casperlin.ee09@nycu.edu.tw; jerryhuang@nycu.edu.tw).}
\thanks{H.-Y.~Lin is with Simula UiB, N--5006 Bergen, Norway (email: lin@simula.no).}
\thanks{\hy Y.-P.~Hsu is with the Department of Communication Engineering, National Taipei University, Taiwan (email: yupinhsu@mail.ntpu.edu.tw).}
}}
{\author{\IEEEauthorblockN{Kuan-Yu Lin\IEEEauthorrefmark{1}, Hsuan-Yin Lin\IEEEauthorrefmark{2}, Yu-Pin Hsu\IEEEauthorrefmark{3}, and Yu-Chih Huang\IEEEauthorrefmark{1}}
  \IEEEauthorblockA{\IEEEauthorrefmark{1}%
    Institute of Communications Engineering, National Yang Ming Chiao Tung University}
  \IEEEauthorblockA{\IEEEauthorrefmark{2}Simula UiB, N--5006 Bergen, Norway}
  \IEEEauthorblockA{\IEEEauthorrefmark{3}%
    Department of Communication Engineering, National Taipei University}\thanks{This work was supported by the National Science and Technology Council of Taiwan under grants MOST~111--2221--E--A49--069--MY3 and 110--2221--E--305--008--MY3.}}
%
}


\maketitle


\begin{abstract}
This paper explores differentially-private federated learning (FL) across time-varying databases, delving into a nuanced three-way tradeoff involving age, accuracy, and differential privacy (DP). Emphasizing the potential advantages of scheduling, we propose an optimization problem aimed at meeting DP requirements while minimizing the loss difference between the aggregated model and the model obtained without DP constraints. To harness the benefits of scheduling, we introduce an age-dependent upper bound on the loss, leading to the development of an age-aware scheduling design. Simulation results underscore the superior performance of our proposed scheme compared to FL with classic DP, which does not consider scheduling as a design factor. This research contributes insights into the interplay of age, accuracy, and DP in FL, with practical implications for scheduling strategies.
\end{abstract}


\section{Introduction}
\label{sec:intro}



Federated learning (FL) has emerged as a pivotal paradigm in machine learning, revolutionizing the way models are trained across decentralized data sources~\cite{McMahanMooreRamageHampsonArcas17_1, Konecy-etal16_1, LiSahuTalwalkarSmith20_1}. In an FL system, data are locally acquired and processed at each client, and then updated machine learning parameters are sent to a central server for aggregation. A major advantage of FL is that it enables local training without the exchange of personal data between the server and the clients. That offers a robust defense against potential eavesdropping by malicious actors. However, the risk of unintentional disclosure of private information remains, primarily through the analysis of variations in the parameters uploaded by clients. Thus, the application of differential privacy (DP) (or local DP)~\cite{Dwork06_1, DworkMcSherryNissimSmith06_1, GeyerKleinNabi17_1sub} in FL systems has received substantial attention (see, e.g.,~\cite{SeifTandonLi20_1, Wei-etal20_1, YangFanYu20_1, KimGuenlueSchaefer21_1}).

An often-overlooked aspect in most of the existing DP research is the impact of data freshness, especially concerning time-varying databases. To address this gap, recent research in~\cite{ZhangWeiBerryHuang24_1} has introduced the concept of \emph{age-dependent DP}. Their innovative approach recognizes the potential of aging (by the use of stale data inputs and/or the delay in releasing outputs) as a novel strategy for enhancing privacy. This method complements the conventional noise injection techniques typically employed in DP frameworks, providing an alternative perspective for privacy preservation.

The present study investigates the application of DP in FL systems operating with time-varying databases. In such systems, delaying updates from a client can enhance its privacy, but this may come at the cost of reduced accuracy in the trained FL model. Therefore, we build upon the research presented in~\cite{ZhangWeiBerryHuang24_1}, further exploring how the aging of data impacts both privacy and the accuracy of the trained FL model. Our objective is to highlight the potential benefit of scheduling in FL under a DP constraint. To this end, we develop a scheduling policy that governs the timing of data collection and the level of noise injection for each client. The primary aim of this scheduling design is to minimize accuracy loss  difference in the trained FL model while ensuring compliance with pre-established privacy constraints.

Some prior research has explored the concept of data freshness in FL systems. For example, \cite{BuyukatesUlukus21_1} studied a FL problem involving fast-changing datasets. An FL protocol was developed where uplink and downlink updates are triggered only when specific numbers of clients are available and reported, respectively. The study characterized the average age of information for the considered protocol and optimized the amounts of clients it needed to wait. In~\cite{MarfoqNegliaKameniVidal23_1}, Marfog \emph{et al.} studied FL for data streams. In this scenario, data arrives sequentially during the training process, and each client can only store a portion of it due to limited memory. A modified stochastic gradient descent was proposed that assigns each sample a weight based on its observed time. However, these prior studies did not tackle the privacy concern and did not investigate the impact of scheduling. 

\begin{figure*}[t!]
  \centering
  \tikzstyle{descript} = [text = black,align=center, minimum height=1.15cm, align=center, outer sep=0pt,font = \footnotesize]
\tikzstyle{activity} =[align=center,outer sep=1pt]
\begin{tikzpicture}[very thick, black, scale=0.85]
    \small
    \coordinate (O) at (-6,0);
    \coordinate (tc1) at (-5,0);
    \coordinate (tc1nextslot) at (-2,0);
    \coordinate (tc3) at (1,0);
    \coordinate (tc2) at (4,0);
    \coordinate (tagg) at (7,0);
    \coordinate (taggnextslot) at (10,0);
    \coordinate (F) at (11,0);
    
    \fill[color=Plum!30] rectangle (tc1) -- (tc1nextslot) -- ($(tc1nextslot)+(0,1.7)$) -- ($(tc1)+(0,1.7)$);
    \fill[color=RedOrange!20] rectangle (tc3) -- (tc2) -- ($(tc2)+(0,1.7)$) -- ($(tc3)+(0,1.7)$);
    \fill[color=Plum!10] rectangle (tc2) -- (tagg) -- ($(tagg)+(0,1.7)$) -- ($(tc2)+(0,1.7)$);
    \fill[color=MidnightBlue!20] rectangle (tagg) -- (taggnextslot) -- ($(taggnextslot)+(0,1.7)$) -- ($(tagg)+(0,1.7)$);
    
    \draw ($(tc1)+(1.5,1.3)$) node[activity] {\footnotesize Client 1:};
    \draw ($(tc1)+(1.5,0.8)$) node[activity] {\footnotesize Data collection};
    \draw ($(tc1)+(1.5,0.3)$) node[activity] {\footnotesize and training};
    
    \draw ($(tc3)+(1.5,1.3)$) node[activity] {\footnotesize Client 3:};
    \draw ($(tc3)+(1.5,0.8)$) node[activity] {\footnotesize Data collection};
    \draw ($(tc3)+(1.5,0.3)$) node[activity] {\footnotesize and training};
    
    \draw ($(tc2)+(1.5,1.3)$) node[activity] {\footnotesize Client 2:};
    \draw ($(tc2)+(1.5,0.8)$) node[activity] {\footnotesize Data collection};
    \draw ($(tc2)+(1.5,0.3)$) node[activity] {\footnotesize and training};
    
    \draw ($(tagg)+(1.5,1.1)$) node[activity] {\footnotesize Parameter Server:};
    \draw ($(tagg)+(1.55,0.6)$) node[activity] {\footnotesize Weight aggregation};
    
    \draw[->] (O) -- (F);
    \foreach \x in {-5,-2,...,11}
    \draw(\x cm,3pt) -- (\x cm,-3pt);
    \foreach \i \j in {-5/$t_{\textnormal{c},1}$,1/$t_{\textnormal{c},3}$,4/$t_{\textnormal{c},2}$,7/$t_{\textnormal{agg}}$}{
      \draw (\i,0) node[below=3pt] {\j} ;
      
      \node[descript,fill=Plum!25,text=Plum](weightuploadblock) at ($(tc1nextslot)+(0,2.6)$) {
        \begin{minipage}{0.21\textwidth}
          \centering
          {\textcolor{black}{\scriptsize Training weight upload\\ based on $t_{\textnormal{c},1}$ collecting time}}
        \end{minipage}};
      \node[descript,fill=RedOrange!15,text=Plum](weightuploadblock) at ($(tc2)+(-0.6,2.6)$) {
        \begin{minipage}{0.21\textwidth}
          \centering
          {\textcolor{black}{\scriptsize Training weight upload\\ based on $t_{\textnormal{c},3}$ collecting time}}
        \end{minipage}};
      \node[descript,fill=Plum!10,text=Plum](weightuploadblock) at ($(tagg)+(1,2.6)$) {
        \begin{minipage}{0.21\textwidth}
          \centering
          {\textcolor{black}{\scriptsize Training weight upload\\ based on $t_{\textnormal{c},2}$ collecting time}}
        \end{minipage}};
      
      \draw[->, color=blue, line width=2pt] ($(tc1nextslot)+(0,0)$) -- ($(tc1nextslot)+(0,2)$);
      \draw[->, color=blue, line width=2pt] ($(tc2)+(0,0)$) -- ($(tc2)+(0,2)$);
      \draw[->, color=blue, line width=2pt] ($(tagg)+(0,0)$) -- ($(tagg)+(0,2)$);
      
      
    }
\end{tikzpicture}
  \caption{ Illustration of the time-varying FL procedures for the case of three clients. $t_{\textnormal{c},i}$ denotes the scheduled time to collect data for each client, and $t_\textnormal{agg}$ is the aggregation time by the central server.}
  \label{fig:FL_procedure}
\end{figure*}
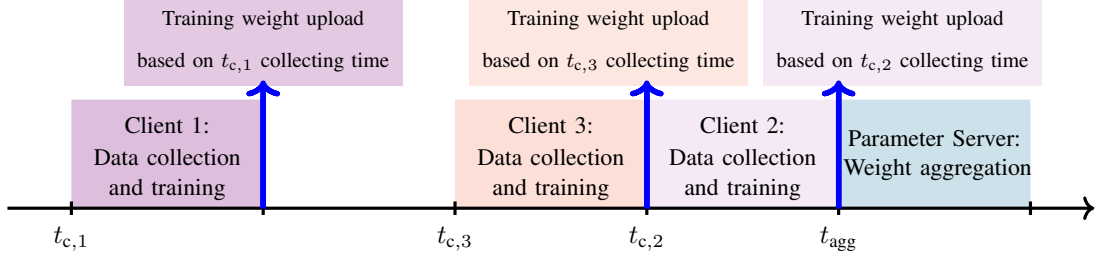

To address the aforementioned tradeoff, in this paper, we consider a privacy-sensitive FL model in a time-varying system. The data collected by each client at different times may be different, following a Markov process that varies from client to client. Moreover, the server is well aware of the varying behaviors of clients' data. The general FL model that we have considered covers the conventional version in which all procedures are assumed to be finished at the same time. See the illustration of our time-varying FL model in Fig.~\ref{fig:FL_procedure}. In the literature, it is known that privacy can be provided by data aging~\cite{ZhangWeiBerryHuang24_1} and adding artificial noise~\cite{Dwork06_1, DworkMcSherryNissimSmith06_1, GeyerKleinNabi17_1sub}, each approach having a different impact on prediction accuracy, hinting at the benefit of scheduling.


\ifthenelse{\boolean{arXiv}}{
\begin{figure*}[t!]
  \centering
  \begin{subfigure}{0.3\linewidth}
    \centering
%
%
\definecolor{mycolor1}{rgb}{0.00000,0.44700,0.74100}%
\begin{tikzpicture}[scale=0.28]

\begin{axis}[%
width=6.00in,
height=5.5in,
at={(1.011in,0.642in)},
scale only axis,
xmin=1,
xmax=10,
xlabel style={font=\color{white!15!black}, yshift=-3mm},
xlabel={Data collection time},
label style = {font=\Huge},
tick label style={font=\Large},
ymode=log,
ymin=727.115553656492,
ymax=1505.94622565766,
yminorticks=true,
ylabel style={font=\color{white!15!black},yshift=3mm},
ylabel={Loss difference},
label style = {font=\Huge},
tick label style={font=\Large},
axis background/.style={fill=white},
xmajorgrids,
ymajorgrids,
yminorgrids
]
\addplot [color=mycolor1, mark=o, mark options={solid, mycolor1}, forget plot]
  table[row sep=crcr]{%
1	1505.94622565766\\
2	1409.04344822553\\
3	1311.1506094302\\
4	1215.40545171686\\
5	1121.33384803513\\
6	1032.13663116542\\
7	945.344089982712\\
8	865.110784546251\\
9	792.23779050376\\
10	727.115553656492\\
};
\end{axis}
\end{tikzpicture}%
    \subcaption{Client 1}
    \label{fig:loss_u1}    
  \end{subfigure}
  \hfill  
  \begin{subfigure}{0.3\linewidth}
    \centering
%
%
\definecolor{mycolor1}{rgb}{0.00000,0.44700,0.74100}%
\begin{tikzpicture}[scale=0.28]

\begin{axis}[%
width=6.00in,
height=5.50in,
at={(1.011in,0.642in)},
scale only axis,
xmin=1,
xmax=10,
xlabel style={font=\color{white!15!black}, yshift=-3mm},
xlabel={Data collection time},
label style = {font=\Huge},
tick label style={font=\Large},
ymode=log,
ymin=257.472334225227,
ymax=752.368115582738,
yminorticks=true,
ylabel style={font=\color{white!15!black},yshift=3mm},
ylabel={Loss difference},
label style = {font=\Huge},
tick label style={font=\Large},
xmajorgrids,
ymajorgrids,
yminorgrids
]
\addplot [color=mycolor1, mark=o, mark options={solid, mycolor1}, forget plot]
  table[row sep=crcr]{%
1	266.693698142604\\
2	257.472334225227\\
3	259.499593522047\\
4	277.230730319675\\
5	318.874343843319\\
6	392.490373930807\\
7	502.801206690064\\
8	640.489389818098\\
9	752.368115582738\\
10	725.672446133692\\
};
\end{axis}
\end{tikzpicture}%
    \subcaption{Client 2}
    \label{fig:loss_u2}    
  \end{subfigure}
  \hfill
  \begin{subfigure}{0.3\linewidth}
    \centering
%
%
\definecolor{mycolor1}{rgb}{0.00000,0.44700,0.74100}%
\begin{tikzpicture}[scale=0.28]

\begin{axis}[%
width=6.00in,
height=5.50in,
at={(1.011in,0.642in)},
scale only axis,
xmin=1,
xmax=10,
xlabel style={font=\color{white!15!black}, yshift=-3mm},
xlabel={Data collection time},
label style = {font=\Huge},
tick label style={font=\Large},
ymode=log,
ymin=341.477077275506,
ymax=725.145508323089,
yminorticks=true,
ylabel style={font=\color{white!15!black},yshift=3mm},
ylabel={Loss difference},
label style = {font=\Huge},
tick label style={font=\Large},
axis background/.style={fill=white},
xmajorgrids,
ymajorgrids,
yminorgrids
]
\addplot [color=mycolor1, mark=o, mark options={solid, mycolor1}, forget plot]
  table[row sep=crcr]{%
1	427.521046286498\\
2	449.996706071094\\
3	464.142513632761\\
4	444.73805197944\\
5	386.98641054604\\
6	341.477077275506\\
7	370.382917916545\\
8	521.729042134117\\
9	708.403887841562\\
10	725.145508323089\\
};
\end{axis}
\end{tikzpicture}%
    \subcaption{Client 3}
    \label{fig:loss_u3}    
  \end{subfigure}
  \caption{{\hy Accuracy loss difference versus data collection time. In each subfigure, we plot the accuracy loss versus the data collection time of a particular client, while fixing that of the other two clients to be the freshest time. For the simulations, relative DP noises are added to ensure a specific privacy level for each client. The simulation settings will be detailed in Section~\ref{sec:discussion_simulation}.}}
  \label{fig:loss_gap_versus_collect_time}
\end{figure*}
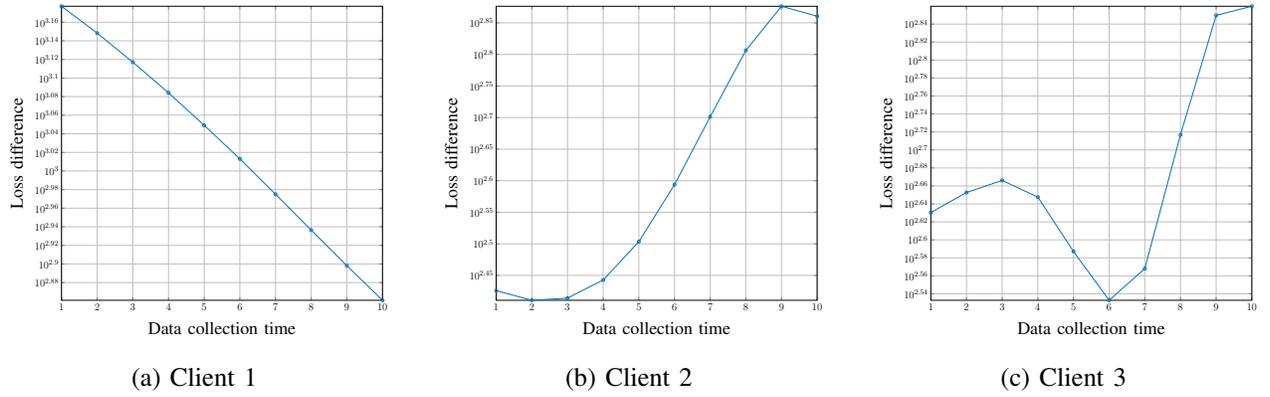}{}

This study marks the first attempt to examine the three-way trade-off between data freshness, privacy, and the accuracy of FL. \ifthenelse{\boolean{arXiv}}{{\hy We provide a three-client toy example to show the phenomenon. Fig.~\ref{fig:loss_gap_versus_collect_time} demonstrates that collecting data at different scheduled times under a specific privacy guarantee for each client may affect the accuracy loss of the global model. Moreover, with the different setting of data varying for each client, the trade-off may also be different. For example, by choosing the freshest collection time of the second and third clients and varying that of Client 1, Fig.~\ref{fig:loss_u1} shows that training the local model with the fresher data and adding the relative DP noise to guarantee each client's privacy level results in less accuracy loss. Alternatively, Fig.~\ref{fig:loss_u2} and Fig.~\ref{fig:loss_u3} show that an optimal scheduled data collection time exists to minimize the accuracy loss under the same privacy guarantee for each client. Thus, neither the freshest data nor the stalest data provide the minimum accuracy loss of the FL global model. The details of the simulation setup are given in Section~\ref{sec:discussion_simulation}.}}{} First, we recognize the transfer learning nature in the considered learning task and extend a recent bound on the generalization error of transfer learning~\cite{WuMantonAickelinZhu20_1} to our model, thereby upper-bounding the loss difference. An optimization problem is then formulated, aiming to find the optimal policy that results in the minimum loss difference while meeting the privacy constraint. Based on our analytical results and those in~\cite{ZhangWeiBerryHuang24_1}, a procedure is developed to find suitable scheduling for clients, who then adaptively adjust the noise levels according to the privacy requirement and the ages of the data. Simulation results show that the proposed policy achieves a better accuracy-privacy trade-off than benchmark schemes that perform random scheduling and/or non-adaptive noise.

The rest of the paper is organized as follows: In Section~\ref{sec:preliminaries}, we discuss preliminaries, including notational conventions, the system model, age-dependent Differential Privacy (DP), and the formulated problem. In Section~\ref{sec:main_result}, we present our analysis of the loss difference and the proposed FL protocol. Simulation results, along with some discussion, are provided in Section~\ref{sec:discussion_simulation}, followed by the conclusion in Section~\ref{sec:conclusion}.



\section{Preliminaries}
\label{sec:preliminaries}

\subsection{Notation}
\label{sec:notation}
Throughout the paper, constants and random variables (RVs) are written in lowercase and uppercase, respectively, for example, $x$ and $X$. Sets are written in calligraphic capital letters, for example, $\mathcal{A}$. $X\sim P_X$ denotes that $X$ is distributed according to a probability distribution $P_X$. For a positive integer $a$, $[a]$ is an abbreviation of the set $\{1, 2, \ldots, a\}$. $\norm{\cdot}$ and $\norm[1]{\cdot}$ denotes the normal norm and $1$-norm, respectively. $\MI{\cdot}{\cdot}$ and $\relDf{\cdot}{\cdot}$ represent the mutual information and the Kullback-Leibler (KL) divergence, respectively.

\subsection{System Model}
\label{sec:system-model}
First, we introduce our FL model, which deals with time-varying data in this subsection. 
Next, we describe the threat model and the privacy mechanism considered in this work in Section~\ref{subsec:age-DP}. Finally, we define our newly formulated problem in section \ref{subsec:problem_formulation}.

\subsubsection{Federated Learning for Time-Varying Data}
\label{subsec:FL}

Similar to the conventional FL model, there are a parameter server (PS) and $m$ clients collaboratively training a machine learning model. However, unlike conventional assumptions in FL, we assume that the data from each client $i\in [m]$, are drawn from a Markov process. Specifically, we define the time-varying data of client $i$ to be $\{Z_{i}^{(\tau)}\}_{\tau\in\Naturals}$, $i\in [m]$, which can be captured by a discrete-time stationary process. Here, $Z_{i}^{(t)}$ is the data of client $i$ at time $t$ over a finite state alphabet $\set{Z}_i$. In addition, we denote by $\mu_i^{(t)}$ the generic distribution of the RV $Z_{i}^{(t)}$. Based on the stationarity, the data follows a stationary probability distribution $\pi_i:\set{Z}_i\to [0,1]$, such that $\pi_i(x) = \bigPrv{Z_{i}^{(t)}=x}$. Also, we denote the $t$-step transition probability by $P_{i,t}(x,y)\eqdef\bigPrvcond{{Z_{i}^{(t+t_0)}=y}}{Z_{i}^{(t_0)}=x}$ and $\hat{P}_{i,t}(x,y)$ is the \emph{reverse process} of the $t$-step transition probability~\cite{LevinPeresWilmer17_1}, defined as follows.
\begin{IEEEeqnarray*}{c}
  \hat{P}_{i,t}(x,y)\eqdef\frac{\pi_i(y)P_{i,t}(y,x)}{\pi_i(x)},\quad\forall\,x,y\in\set{Z}_i, t\in\Naturals, i\in [m].
\end{IEEEeqnarray*}

The PS aims to achieve a well-trained (data-driven) global model at a certain aggression time $t_{\textnormal{agg}}$. 
To achieve this goal, the server first schedules the $i$-th client to perform the data collection procedure at time $t_{\textnormal{c},i}$, where $\tc{i}\in [t_{\textnormal{agg}}]$, then waits for the client to do local training and upload their training parameters. For energy-saving purposes, we assume that the clients do not need to be active all the time. The clients are only active with their initial states at time $\tc{i}$, and start collecting data and training immediately. 
After all participating clients finish uploading the parameters, the server aggregates them from all clients and achieves the global parameters $\vect{w}$ at time $t_{\textnormal{agg}}$, which is expressed as
\begin{IEEEeqnarray*}{c}
  \vect{w}(\set{T}_\txtc)\eqdef\sum_{i=1}^m p_i\vect{w}_i^{(\tc{i})},
\end{IEEEeqnarray*}
where $\vect{w}_i^{(\tc{i})}$ is the local parameter trained by the client $i$ at time $\tc{i}$, $i\in [m]$, $\set{T}_\textnormal{c}\eqdef\{t_{\textnormal{c},i}\}_{i\in [m]}$, $p_i\eqdef\nicefrac{\ecard{\set{D}_i^{(t_{\textnormal{c},i})}}}{\ecard{\set{D}}} \geq 0$ with $\ecard{\set{D}}=\sum_{i=1}^m\ecard{\set{D}_i^{(t_{\textnormal{c},i})}}$, and $\set{D}_i^{(\tc{i})}$ is the training dataset in which each data is independent and identically distributed (i.i.d.) sampled from $\mu_i^{(t_{c,i})}$. Here, we also assume that $\ecard{\set{D}_i^{(t_{\textnormal{c},i})}}=\ecard{\set{D}_i}$, for all $t_{\textnormal{c},i}\in [t_{\textnormal{agg}}]$.


\subsubsection{Empirical risk minimization}
\label{sec:empir-risk-minim}

In this work, we consider the \emph{empirical risk minimization (ERM)} formulation. The ERM solution $\hat{\vect{w}}$ with respect to the time-vary dataset at the aggression time $\tagg$ can be derived by the following formula.
\begin{IEEEeqnarray*}{c}
  \hat{\vect{w}} = \argmin_{\vect{w}}\biggl[\hat{\Loss}(\vect{w},\set{D}^{(\tagg)})\eqdef\sum_{i=1}^m p_i\hat{\Loss}_{i}(\vect{w},\set{D}_i^{(\tagg)})\biggr],
\end{IEEEeqnarray*}
where $\hat{\Loss}_i(w,\set{D}_i^{(t)})\eqdef\frac{1}{\ecard{\set{D}_i^{(t)}}}\sum_{j=1}^{\ecard{\set{D}^{(t)}_i}}\ell(w,Z^{(t)}_{i,j})$ is an empirical loss function for a hypothesis $w\in\set{W}$ over the training dataset $\set{D}^{(t)}_i=\{Z_{i,1}^{(t)},\ldots,Z_{i,\ecard{\set{D}_i}}^{(t)}\}$, $Z^{(t)}_{i,j}\sim\mu_i^{(t)}$, $\forall\,j\in [\ecard{\set{D}_i}]$, $\set{D}^{(t)} = \{\set{D}_i^{(t)}\}_{i\in [m]}$, and $\ell(\cdot,\cdot)$ is a certain loss function. With the goal to determine the best $\vect{w}$ that minimizes the \emph{population risk} with respect to the set of all target distributions of the clients at time $\tagg$, we consider the average loss $\avgLoss(\vect{w})\eqdef\sum_{i=1}^m p_i\Loss_{\mu_i}(w)$, where $\Loss_{\mu_i}(w)\eqdef\bigE[Z_i\sim\mu_i]{\ell(\vect{w},Z_i)}$, and $\mu_i$ represents the shorthand of $\mu_i^{(\tagg)}$.

\subsection{Age-dependent Differential Privacy}
\label{subsec:age-DP}

We assume the server is honest, but there exist external adversaries who may have access to $\vect{w}(\set{T}_\txtc)$ and aim to obtain sensitive information from the clients through $\vect{w}(\set{T}_\txtc)$. One approach to ensure privacy is to employ classical differential privacy (DP) mechanisms on both each client's update $\vect{w}_i^{(t_{\textrm{c},i})}$ and the aggregated model $\vect{w}(\set{T}_\txtc)$~\cite{Wei-etal20_1}. In this work, we choose to implement DP mechanisms solely on the client side for simplicity and explore both classic DP and age-dependent DP mechanisms.

We first review the \emph{classical} $\eps$-DP to provide a privacy guarantee for each client.
\begin{definition}[{Classical $\eps$-Differential Privacy ($\eps$-DP)}]
\label{def:Pure-DP}
  A randomized mechanism $M: \set{D}\to\set{Y}$ is called $\epsilon_{\textnormal{c}}$-DP if for every pair of datasets $\set{D},\set{D}'$ which differ in only one data, we have
  \begin{IEEEeqnarray*}{c}
    \bigPrv{M(\set{D})\in\set{S}}\leq \exp(\epsilon_{\textnormal{c}}) \bigPrv{M(\set{D}')\in\set{S}},\,\forall\,\set{S}\subseteq\set{Y},\IEEEeqnarraynumspace\label{eqn:classical-DP}
  \end{IEEEeqnarray*}
  where the probability is taken over the randomness of mechanism $M$.
\end{definition}
In addition to the classical $\eps$-DP, the aging of the data also provides another domain of privacy guarantee. Such a concept is referred to as \emph{age-dependent DP} and was proposed by~\cite{ZhangWeiBerryHuang24_1}.

\begin{definition}[{Age-Dependent DP~\cite[Def.~4]{ZhangWeiBerryHuang24_1}}]
  \label{def:Age-DP}
  A randomized mechanism $M:\set{D}\to\set{Y}$ is $(\epsilon(t), t)$-age-dependent DP for a given random process $\{Z^{(\tau)}\}_{\tau\in\Naturals}$ if for every pair of datasets $\set{D}$, $\set{D}'$ which differ in only one data, we have
  \begin{IEEEeqnarray*}{rCl}
    \IEEEeqnarraymulticol{3}{l}{%
      \bigPrvcond{M(Z^{(t_0)})\in\set{S}}{Z^{(t_0+t)}=\set{D}}}
    \nonumber\\
    & \leq &\exp{\bigl(\epsilon(t)\bigr)}\bigPrvcond{M(Z^{(t_0)})\in\set{S}}{Z^{(t_0+t)}=\set{D}'},\,\forall\,\set{S}\subseteq\set{Y},\IEEEeqnarraynumspace\label{eqn:age-DP}
  \end{IEEEeqnarray*}
  for any $t_0\in\Naturals$, where the probability takes into account the randomness of the output $M$ and $\{Z^{(\tau)}\}_{\tau\in\Naturals}$.
\end{definition}
Def.~\ref{def:Age-DP} reduces to Def.~\ref{def:Pure-DP} when $t=0$. The main difference between Defs.~\ref{def:Pure-DP} and~\ref{def:Age-DP} is that the classical DP does not consider any information about the data distribution.

The $\epsilon(t)$ in Definition~\ref{def:Age-DP} relates to the aging and classical-DP. We have the following proposition from~\cite[Th.~1]{ZhangWeiBerryHuang24_1}.
\begin{proposition}[Mechanism-Dependent Guarantee]
  \label{prop:eps_AoI-DP_TotalVariationDistance}
Consider an $\epsilon_{\textnormal{c}}$-DP mechanism for the $i$-th client, $i\in[m]$. Then, it is also $\epsilon(t,{\epsilon_\textnormal{c}})$-age-dependent DP, where $\epsilon(t,\eps_{\textnormal{c}})$ is defined as 
\begin{IEEEeqnarray}{c}
  \epsilon(t,\eps_{\textnormal{c}})\eqdef\ln\bigl[1+\Delta_i(t)\bigl(\exp{(\epsilon_{\textnormal{c}})} - 1\bigr)\bigr],\quad\forall\,t\in\Naturals,\label{eqn:expression_epsilon_t}
\end{IEEEeqnarray}
where $\Delta_i(t)$ is the \emph{total variation distance} for the $i$-th client, given by
\begin{IEEEeqnarray*}{c}
  \Delta_i(t)\eqdef
  \max_{x_{i},x'_{i}\in\set{Z}_i}\bigdTV{\hat{P}_{i,t}(x_{i},\cdot)}{\hat{P}_{i,t}(x'_{i},\cdot)},\,\forall\,t\in\Naturals,\IEEEeqnarraynumspace\label{eqn:def_Delta_t}
\end{IEEEeqnarray*}
and $\edTV{\cdot}{\cdot}$ is the total variation distance between probability distributions $\mu$ and $\nu$ on a finite set $\set{F}$:
\begin{IEEEeqnarray*}{c}
  \edTV{\mu}{\nu}\eqdef\max_{\mathcal{A}\subset\set{F}}\bigcard{\mu(\mathcal{A}) - \nu(\mathcal{A})}.
  \label{eqn:def_dTV}
\end{IEEEeqnarray*}
\end{proposition}

\subsection{Problem Formulation}
\label{subsec:problem_formulation}

As we mentioned in Section~\ref{sec:intro}, the PS should first schedule the clients when to perform data collection and upload their parameters in order to deploy the well-train model at a specific time. Moreover, we know that using different data times can provide different privacy levels but bring penalties to the accuracy of the global model. In this paper, we first characterize the accuracy loss difference between the global model's accuracy loss, which is governed by a set of specific scheduling times $\set{T}_\txtc$ with the corresponding age-dependent DP mechanisms, and that of the optimal well-trained model, which is the model trained by the freshest data without applying any privacy mechanism. 
More specifically, we formulate an optimization problem to find the scheduling times to do data collection for each client and the corresponding noise to minimize the accuracy loss difference under a certain privacy guarantee $\epsilon(t_{\textnormal{c},i},\eps_{\textnormal{c},i})\leq\bar{\eps}$, $\forall\,i\in [m]$, which is expressed as follows,
\begin{IEEEeqnarray}{rCl}
  \IEEEyesnumber\label{eqn:optimization_ERM}
  \IEEEyessubnumber*
  \min_{\set{T}_\textnormal{c},\set{E}_\textnormal{c}}&&\,\,\BigE{\hat{\Loss}\bigl(\tilde{\vect{W}}(\set{T}_\textnormal{c},\set{E}_\textnormal{c}),\set{D}^{(t_{\textnormal{agg}})}\bigr)-\hat{\Loss}(\vect{W}^{(\tagg)},\set{D}^{(t_{\textnormal{agg}})})},\IEEEeqnarraynumspace\label{eqn:def_loss_gap}
  \\
  \textnormal{subject to}& &\qquad\epsilon({\hy \tagg-\tc{i}},\eps_{\textnormal{c},i})\leq\bar{\epsilon},\quad\forall\,i\in [m],\label{eqn:privacy_constraints}
\end{IEEEeqnarray}
where the expectation is taken over the distribution $P_{W(\set{T}_\txtc),N,W^{(\tagg)}}$, $i\in [m]$, $\set{E}_{\textnormal{c}}\eqdef\{\eps_{\textnormal{c},i}\}_{i\in [m]}$, $\tilde{\vect{W}}(\set{T}_{\textnormal{c}},\set{E}_\textnormal{c})=\vect{W}(\set{T}_\textnormal{c})+N\eqdef\sum_{i=1}^m p_i\bigl(\vect{W}_i^{(\tc{i})}+N_i\bigr)$, $N_i$ denotes the Laplace noise RV added to each client to satisfy $\eps_{\txtc,i}$-DP with $\eps_{\textnormal{c},i}\geq 0$, $i\in [m]$, $W^{(\tagg)}\eqdef\sum_{i=1}^m p_i W_i^{(\tagg)}$, and $\bar{\epsilon}$ is the global target privacy level.


\section{Main Results}
\label{sec:main_result}

Towards solving the problem stated above, in Section~\ref{subsec:Loss_analysis}, we first prove an upper bound on the loss difference caused by the age of information. Based on this bound, we then present the proposed scheduling algorithm in Section~\ref{subsec:Proposed_scheduling}.

\subsection{Loss Difference Analysis}
\label{subsec:Loss_analysis}

Here, we analyze the difference between the empirical loss of $\tilde{\vect{W}}(\set{T}_{\textnormal{c}},\set{E}_\textnormal{c})$ with respect to $\set{D}^{(\tagg)}$ under a certain privacy guarantee, i.e., $\hat{\Loss}\bigl(\tilde{\vect{W}}(\set{T}_{\textnormal{c}},\set{E}_\textnormal{c}),\set{D}^{(\tagg)}\bigr)$, and that of $\vect{W}^{(\tagg)}$ with respect to $\set{D}^{(t_{\textnormal{agg}})}$, where $\tilde{\vect{W}}(\set{T}_{\textnormal{c}},\set{E}_\textnormal{c})$ is the aggregated model obtained by training with every aged datasets $\set{D}_i^{(\tc{i})}$, $i\in [m]$, such that each client satisfies $\eps_{\txtc,i}$-DP, and $\vect{W}^{(\tagg)}$ is trained with the latest dataset $\set{D}^{(t_{\textnormal{agg}})}$ when computing the empirical loss. Thus, it is inherently a transfer learning problem. However, due to the FL nature of the considered problem---where we aggregate only trained models instead of datasets---existing bounds, such as those in~\cite{WuMantonAickelinZhu20_1}, are not directly applicable. In the following, we present the main analytical result. Note that for simplicity of notation, we will sometimes omit the arguments of $\tilde{\vect{W}}(\set{T}_{\textnormal{c}},\set{E}_\textnormal{c})$ and $\vect{W}(\set{T}_{\textnormal{c}})$.

\begin{theorem}[Loss Difference Between Global Practical Weight under Privacy Guarantee and Ideal Weight]
  \label{thm:UB_convergence}
  Let $\tilde{\vect{W}}(\set{T}_\textnormal{c},\set{E}_\textnormal{c})=\vect{W}(\set{T}_\txtc) + N\eqdef\sum_{i=1}^m p_i\bigl(\vect{W}_i^{(\tc{i})}+N_i\bigr)$, where $N_i$ denotes the Laplace noise RV added to each client to satisfy $\eps_{i,\txtc}$-DP, $i\in [m]$, and $W^{(\tagg)}\eqdef\sum_{i=1}^m p_i W_i^{(\tagg)}$. Assume that the hypotheses $W(\set{T}_\txtc) = \sum_{i=1}^m p_i\vect{W}_i^{(\tc{i})}\in\set{W}$ and $W^{(\tagg)}\in\set{W}$ are distributed over $P_W$ and $P_{W^{(\tagg)}}$, respectively. Moreover, the cumulant generating function of the RV, $\ell\bigl(\vect{W}+N,Z_i^{(t_\textnormal{agg})}\bigr) -\bigE{\ell(\vect{W}+N,Z_i^{(t_\textnormal{agg})})}$, is upper bounded by some function $\psi_i(\lambda)$ in the interval $(b_{i,-},b_{i,+})$ under the product distribution $P_W \otimes P_{N}\otimes\mu_i$ with some $b_{i,-}<0$ and $b_{i,+}>0$, $i\in [m]$, and the cumulant generating function of the RV, $\ell\bigl(\vect{W}^{(\tagg)},Z_i^{(t_\textnormal{agg})}\bigr) -\bigE{\ell(\vect{W}^{(\tagg)},Z_i^{(t_\textnormal{agg})})}$, is upper bounded by some function $\phi(\lambda)$ in the interval $(c_{i,-},c_{i,+})$ under the product distribution $P_{W^{(\tagg)}}\otimes P_{N}\otimes\mu_i$ with some $c_{i,-}<0$ and $c_{i,+}>0$. Then, the expected loss difference is bounded from above by
\ifthenelse{\boolean{arXiv}}{
\begin{IEEEeqnarray}{rCl}
  \bigE{\hat{\Loss}\bigl(\tilde{\vect{W}}(\set{T}_\textnormal{c},\set{E}_\textnormal{c}),\set{D}^{(\tagg)}\bigr)-\hat{\Loss}(\vect{W}^{(\tagg)},\set{D}^{(\tagg)})}
  & \leq &\sum_{i=1}^m p_i \frac{1}{\ecard{\set{D}_i}}\sum_{j=1}^{\ecard{\set{D}_i}}\Bigl[\psi_{i,+}^{*-1}\bigl(\bigMI{\vect{W}(\set{T}_\textnormal{c})}{Z_{i,j}^{(\tagg)}}\bigr)
  \nonumber\\
  &&\hspace*{-5mm}+\>\phi_{i,-}^{*-1}\bigl(\bigMI{\vect{W}^{(\tagg)}}{Z_{i,j}^{(\tagg)}}\bigr)\Bigr]
  +\bigE{\Risk(\tilde{\vect{W}}(\set{T}_\textnormal{c},\set{E}_\textnormal{c}),\vect{W}^{(\tagg)})},
  \IEEEeqnarraynumspace\label{eqn:UB_loss}
\end{IEEEeqnarray}}{
\begin{IEEEeqnarray}{rCl}
  \IEEEeqnarraymulticol{3}{l}{%
    \bigE{\hat{\Loss}(\tilde{\vect{W}}(\set{T}_\txtc,\set{E}_\txtc),\set{D}^{(\tagg)})-\hat{\Loss}(\vect{W}^{(\tagg)},\set{D}^{(\tagg)})}  
  }\nonumber\\*\quad
  & \leq &\sum_{i=1}^m p_i\frac{1}{\ecard{\set{D}_i}}\sum_{j=1}^{\ecard{\set{D}_i}}\Bigl[\psi_{i,+}^{*-1}\bigl(\bigMI{\vect{W}(\set{T}_\textnormal{c})}{Z_{i,j}^{(\tagg)}}\bigr)
  \nonumber\\
  &&\hspace*{2.25cm}+\>\phi_{i,-}^{*-1}\bigl(\bigMI{\vect{W}^{(\tagg)}}{Z_{i,j}^{(\tagg)}}\bigr)\Bigr]
  \nonumber\\
  &&\hspace*{2.25cm}+\>\bigE{\Risk\bigl(\tilde{\vect{W}}(\set{T}_\textnormal{c},\set{E}_\textnormal{c}),\vect{W}^{(\tagg)}\bigr)},
  \IEEEeqnarraynumspace\label{eqn:UB_loss}
\end{IEEEeqnarray}
}
where $\ecard{\set{D}_i}=\ecard{\set{D}_i^{(\tc{i})}}=\ecard{\set{D}_{i}^{(\tagg)}}$, $Z^{(\tagg)}_{i,j}\sim\mu_i$, $\forall\,j\in[\ecard{\set{D}_i}]$, $\set{T}_\textnormal{c}=\{\tc{i}\}_{i\in [m]}, \set{E}_{\textnormal{c}} = \{\eps_{\textnormal{c},i}\}_{i\in [m]}$,
\ifthenelse{\boolean{arXiv}}{
\begin{IEEEeqnarray*}{rCl}
  \psi^{*-1}_{i,+}(x)& \eqdef &\inf_{\lambda \in [0,b_{i,+}]} \frac{x+\psi_i(\lambda)}{\lambda},\quad
  \phi^{*-1}_{i,-}(x)\eqdef\inf_{\lambda \in [0,-c_{i,-}]}\frac{x+\phi_i(-\lambda)}{\lambda},
  \\
  \Risk(\vect{W})& \eqdef &\avgLoss(\tilde{\vect{W}})-\avgLoss(\vect{W}^{(\tagg)}).\IEEEeqnarraynumspace
\end{IEEEeqnarray*}}{
\begin{IEEEeqnarray*}{rCl}
  \psi^{*-1}_{i,+}(x)& \eqdef &\inf_{\lambda \in [0,b_{i,+}]} \frac{x+\psi_i(\lambda)}{\lambda},\quad
  \\
  \phi^{*-1}_{i,-}(x)& \eqdef &\inf_{\lambda \in [0,-c_{i,-}]}\frac{x+\phi_i(-\lambda)}{\lambda},
  \\
  \Risk(\vect{W})& \eqdef &\avgLoss(\tilde{\vect{W}})-\avgLoss(\vect{W}^{(\tagg)}).\IEEEeqnarraynumspace
\end{IEEEeqnarray*}
}
Furthermore, if $\vect{W}+N \in \set{W}$, for every $\vect{W}\in\set{W}$, then the risk term $\bigE{\Risk\bigl(\tilde{\vect{W}}(\set{T}_\txtc,\set{E}_\txtc), \vect{W}^{(\tagg)}\bigr)}$ in~\eqref{eqn:UB_loss} can be further bounded by
\begin{equation*}
  \bigE{\Risk\bigl(\tilde{\vect{W}}(\set{T}_\txtc,\set{E}_\txtc),\vect{W}^{(\tagg)}\bigr)}\leq\sup_{\vect{w},\vect{w}'\in\set{W}}\bigl(\avgLoss(\vect{w}) - \avgLoss(w')\bigr),   
\end{equation*}
and the dependency of $N$ in the upper bound can be removed.
\end{theorem}

\ifthenelse{\boolean{arXiv}}{
\begin{IEEEproof}
  See Appendix~\ref{apx:UB_convergence}.
\end{IEEEproof}}{
\begin{IEEEproof}
See Appendix A in~\cite{LinLinHsuHuang24_1sub}.
\end{IEEEproof}
}

\ifthenelse{\boolean{NOTE}}{
{\hy
\subsection{Mean Estimation for Birth-and-Death Markov Chain}

\begin{theorem}
  \label{thm:loss_difference_closed-form_mean-estimation}
  Assume that $Z_i^{(\tagg)}\sim\mu_{i}^{(t_i)}\equiv\mu_i^{(\tc{i})}\mat{P}^{t_i}_i$, $i\in[m]$, $t_i\in \Naturals$, where $\mat{P}_i$ is the transition probability matrix of a $\ecard{\set{Z}}$-state birth-and-death Markov chain for the $i$-th client and $t_i\eqdef\tagg-\tc{i}$, $i\in [m]$. Consider the mean-square error loss function
  \begin{equation*}
    \ell(w,z) = (w-z)^2,
  \end{equation*}
  and the estimated weight to be the mean of samples:
  \begin{IEEEeqnarray*}{rCl}
    W=\sum_{i=1}^m p_i\frac{1}{n_i}\sum_{j=1}^{n_i}Z_{i,j}^{(t)},\quad t\in\Naturals.
  \end{IEEEeqnarray*}
  Then, the loss difference between global practical weight under privacy guarantee and ideal weight can be expressed as
  \begin{IEEEeqnarray*}{rCl}
    \IEEEeqnarraymulticol{3}{l}{%
    \bigE{\hat{\Loss}(\tilde{\vect{W}}(\set{T}_\txtc,\set{E}_\txtc),\set{D}^{(\tagg)})}-\bigE{\hat{\Loss}(\vect{W}^{(\tagg)},\set{D}^{(\tagg)})}}\nonumber\\*\quad%
  & = &\sum_{i=1}^m 2 p^2_i\left(\frac{z_{i,\textnormal{max}}-z_{i,\textnormal{min}}}{n_i g_i(t_i,\bar{\eps})}\right)^2
  +\sum_{i=1}^m p^2_i\frac{\bigVar{Z_{i}^{(\tc{i})}}}{n_i}+\biggl(\sum_{i=1}^m p_i\bigE{Z_{i}^{(\tc{i})}}\biggr)^2
  \nonumber\\
  && -\>2\sum_{i=1}^m\frac{p_{i}^2}{n^2_i}\Bigl(n_{i}\bigE{Z^{(\tagg)}_{i}Z^{(\tc{i})}_{i}}+n_i(n_i-1)\bigE{Z^{(\tagg)}_{i}}\bigE{Z^{(\tc{i})}_{i}}\Bigr)
  \nonumber\\
  && -\>2\sum_{i=1}^m p_i\bigE{Z^{(\tagg)}_{i}}\sum_{\substack{i'\neq i}}^m p_{i'}\bigE{Z^{(\tc{i'})}_{i'}}
  +\sum_{i=1}^m p^2_i\frac{\bigVar{Z_i^{(\tagg)}}}{n_i}+\biggl(\sum_{i=1}^m p_i\bigE{Z_{i}^{(\tagg)}}\biggr)^2,\IEEEeqnarraynumspace\label{eq:loss-difference_mean-estimation}
  \end{IEEEeqnarray*}
  where 
  \begin{IEEEeqnarray}{c}
    g_i(t_i,\bar{\eps})\eqdef\ln{\Bigl(\frac{\ope^{\bar{\eps}}-1}{\Delta_i(t_i)}+1\Bigr)},
    \label{eq:exact-epsc_client-i}
  \end{IEEEeqnarray}
  $z_{i,\textnormal{max}}\eqdef\max\{x_i\in\set{Z}_i\colon x_i\}$ and $z_{i,\textnormal{min}}\eqdef\min\{x_i\in\set{Z}_i\colon x_i\}$, and $n_i\eqdef\ecard{\set{D}^{(t)}_i}$, $i\in [m]$.
\end{theorem}
\begin{IEEEproof}
  \subsubsection{Privacy Analysis}
  \label{sec:privacy-analysis_AoI-DP}
  
  Recall Proposition~\ref{prop:eps_AoI-DP_TotalVariationDistance}, we can get the relation between $t_i\eqdef\tagg-\tc{i}$, $\epsc{i}$ for the $i$-th client, and a certain privacy guarantee $\bar{\eps}$:
  \begin{IEEEeqnarray*}{c}
    \bar{\eps}=\eps(t_i,\eps_{\txtc,i})=\ln\bigl[1+\Delta_i(t_i)\bigl(\exp{(\epsilon_{\txtc,i})} - 1\bigr)\bigr].
  \end{IEEEeqnarray*}
  Hence, each client $i$ is using an $\eps_{\txtc,i}$-DP mechanism with
  \begin{IEEEeqnarray*}{c}
    \eps_{\txtc,i}(t_i,\bar{\eps})=\ln{\Bigl(\frac{\ope^{\bar{\eps}}-1}{\Delta_i(t_i)}+1\Bigr)}\eqdef g_i(t_i,\bar{\eps}).
  \end{IEEEeqnarray*}
  
Moreover, for the $i$-th client, $i\in [m]$, the released data $M\bigl(Z_i^{(\tc{i})}\bigr)$ is expressed as
\begin{IEEEeqnarray*}{rCl}
  M(Z_i^{(\tc{i})})& = &f(z^{(\tc{i})}_{i})+L_i\eqdef\frac{1}{n_i}\sum_{j=1}^{n_i}Z^{(\tc{i})}_{i,j}+L_i,
\end{IEEEeqnarray*}
where $L_i\sim\Lap{\eta_i}$ is the Laplace noise RV with parameter $\eta_i\geq 0$. The sensitivity of this averaging function $f$ is
\begin{IEEEeqnarray*}{c}
  s_i^{(f)}=\frac{z_{i,\textnormal{max}}-z_{i,\textnormal{min}}}{n_i},
\end{IEEEeqnarray*}
where $z_{i,\textnormal{max}}\eqdef\max\{x_i\in\set{Z}_i\colon x_i\}$ and $z_{i,\textnormal{min}}\eqdef\min\{x_i\in\set{Z}_i\colon x_i\}$. Hence, to achieve $\eps_{\textnormal{c},i}$-DP, one would need to use a Laplace noise RV $L_i\sim\eLap{\eta_i}$ with
\begin{IEEEeqnarray*}{rCl}
  \eta_i& = &\frac{s^{(f)}}{\eps_{\textnormal{c},i}}=\frac{z_{i,\textnormal{max}}-z_{i,\textnormal{min}}}{n_i\eps_{\textnormal{c},i}},
\end{IEEEeqnarray*}
and hence
\begin{IEEEeqnarray}{c}
  \eVar{L_i}=2\eta_i^2=2\left(\frac{z_{i,\textnormal{max}}-z_{i,\textnormal{min}}}{n_i g_i(t_i,\bar{\eps})}\right)^2.\label{eq:variance_Laplace-RV}
\end{IEEEeqnarray}

\subsubsection{Accuracy Analysis}
\label{sec:accuracy-analysis}

Assume that $Z_i^{(\tagg)}\sim\mu_{i}^{(\tagg)}\equiv\mu_i^{(\tc{i})}\mat{P}_i^{t_i}$, $i\in[m]$, $t_i\in \Naturals$, where $\mat{P}_i$ is the transition probability matrix of a $\ecard{\set{Z}}$-state birth-and-death Markov chain for the $i$-th client. We define the loss function to be the mean-square error:
\begin{equation*}
  \ell(w,z) = (w-z)^2.
\end{equation*}

The practical weight under a certain privacy guarantee and ideal weight can be expressed as
\begin{IEEEeqnarray*}{rCl}
  \tilde{\vect{W}}(\set{T}_\textnormal{c},\set{E}_\textnormal{c})& = &\sum_{i=1}^m p_i(W_i^{(\tc{i})}+L_i)=\sum_{i=1}^m p_i\Bigl(\frac{1}{n_i}\sum_{j=1}^{n_i} Z_{i,j}^{(\tc{i})} + L_i\Bigr),
  \\
  \vect{W}^{(\tagg)}& = &\sum_{i=1}^m p_i W_i^{(\tagg)}=\sum_{i=1}^m p_i\Bigl(\frac{1}{n_i}\sum_{j=1}^{n_i} Z_{i,j}^{(\tagg)}\Bigr),
\end{IEEEeqnarray*}
where $n_i\eqdef\ecard{\set{D}^{(t)}_i}$, $Z^{(\tc{i})}_{i}\sim\mu^{(\tc{i})}_i$ is the initial state distribution, and $Z_{i}^{(\tagg)}\sim\mu_i=\mu_i^{(\tc{i})}\mat{P}^{\tagg-\tc{i}}$.

\smallskip

\noindent
{\bf Loss for practical weight:}
The expected empirical loss for $\tilde{W}(\set{T}_\txtc,\set{E}_\txtc)$ is given by
\begin{IEEEeqnarray*}{rCl}
  \bigE{\hat{\Loss}(\tilde{\vect{W}}(\set{T}_\txtc,\set{E}_\txtc),\set{D}^{(\tagg)})}& = &
  \sum_{i=1}^m p_i\BigE{\hat{\Loss}_{i}(\tilde{\vect{W}}(\set{T}_\txtc,\set{E}_\txtc),\set{D}_i^{(\tagg)})}
  \\
  & = &\sum_{i=1}^m p_i\biggE{\frac{1}{n_i}\sum_{j=1}^{n_i}\Bigl(\tilde{W}-Z^{(\tagg)}_{i,j}\Bigr)^2}
  \\
  & = &\sum_{i=1}^m p_i\frac{1}{n_i}\biggE{\sum_{j=1}^{n_i}\tilde{W}^2-2\tilde{W}\sum_{j=1}^{n_i}Z^{(\tagg)}_{i,j}+\sum_{j=1}^{n_i}\bigl(Z^{(\tagg)}_{i,j}\bigr)^2}
  \\
  & = &\sum_{i=1}^m p_i\frac{1}{n_i}\biggl(\sum_{j=1}^{n_i}\bigE{\tilde{W}^2}-2\bigE{\tilde{W}\bigl(n_i W_i^{(\tagg)}\bigr)}+\sum_{j=1}^{n_i}\bigE{\bigl(Z^{(\tagg)}_{i,j}\bigr)^2}\biggr)
  \\
  & = &\sum_{i=1}^m p_i\frac{1}{n_i}\Bigl(n_i\bigE{\tilde{W}^2}-2n_i\bigE{\vect{W}_i^{(\tagg)}\tilde{\vect{W}}}+n_i\bigE{\bigl(Z^{(\tagg)}_{i}\bigr)^2}\Bigr)
  \\
  & = &\sum_{i=1}^m p_i\Bigl(\bigE{\tilde{W}^2}-2\bigE{\vect{W}_i^{(\tagg)}\tilde{\vect{W}}}+\bigE{\bigl(Z^{(\tagg)}_{i}\bigr)^2}\Bigr)
  \\
  & = &\bigE{\tilde{W}^2}-2\biggE{\sum_{i=1}^m p_i\vect{W}_i^{(\tagg)}\tilde{\vect{W}}}+\sum_{i=1}^m p_i\bigE{\bigl(Z^{(\tagg)}_{i}\bigr)^2}
  \\
  & = &\bigE{\tilde{W}^2}-2\bigE{\vect{W}^{(\tagg)}\tilde{\vect{W}}}+\sum_{i=1}^m p_i\bigE{\bigl(Z^{(\tagg)}_{i}\bigr)^2}.\IEEEyesnumber\label{eq:all-terms_practical-weight}
\end{IEEEeqnarray*}

Now, let us first deal with the second term in~\eqref{eq:all-terms_practical-weight}.
\begin{IEEEeqnarray*}{rCl}
  \bigE{\vect{W}^{(\tagg)}\tilde{\vect{W}}}& = &\biggE{\biggl(\sum_{i=1}^m p_{i} W^{(\tagg)}_{i}\biggr)\biggl(\sum_{i'=1}^m p_{i'}\bigl(W^{(\tc{i'})}_{i'}+L_{i'}\bigr)\biggr)}
  \\
  & = &\biggE{\biggl(\sum_{i=1}^m p_{i} W^{(\tagg)}_{i}\biggr)\biggl(\sum_{i'=1}^m p_{i'}W^{(\tc{i'})}_{i'}+\sum_{i'=1}^m p_{i'} L_{i'}\biggr)}
  \\
  & \stackrel{(a)}{=} &\biggE{\biggl(\sum_{i=1}^m p_{i} W^{(\tagg)}_{i}\biggr)\biggl(\sum_{i'=1}^m p_{i'}W^{(\tc{i'})}_{i'}\biggr)}
  \\
  & \stackrel{(b)}{=} &\sum_{i=1}^m p_{i}^2\bigE{W^{(\tagg)}_{i}W^{(\tc{i})}_{i}}+\sum_{i=1}^m p_i\bigE{W^{(\tagg)}_{i}}\sum_{\substack{i'\neq i}}^m p_{i'}\bigE{W^{(\tc{i'})}_{i'}}
  \\
  & = &\sum_{i=1}^m p_{i}^2\bigE{W^{(\tagg)}_{i}W^{(\tc{i})}_{i}}+\sum_{i=1}^m p_i\bigE{Z^{(\tagg)}_{i}}\sum_{\substack{i'\neq i}}^m p_{i'}\bigE{Z^{(\tc{i'})}_{i'}},
  \\
  & = &\sum_{i=1}^m p_{i}^2\biggE{\biggl(\frac{1}{n_{i}}\sum_{j=1}^{n_{i}}Z^{(\tagg)}_{i,j}\biggr)\biggl(\frac{1}{n_{i}}\sum_{j'=1}^{n_{i}}Z^{(\tc{i})}_{i,j'}\biggl)}+\sum_{i=1}^m p_i\bigE{Z^{(\tagg)}_{i}}\sum_{\substack{i'\neq i}}^m p_{i'}\bigE{Z^{(\tc{i'})}_{i'}},
  \\
  & = &\sum_{i=1}^m\frac{p_{i}^2}{n^2_{i}}\sum_{j=1}^{n_{i}}\sum_{j'=1}^{n_{i}}\bigE{Z^{(\tagg)}_{i,j}Z^{(\tc{i})}_{i,j'}}+\sum_{i=1}^m p_i\bigE{Z^{(\tagg)}_{i}}\sum_{\substack{i'\neq i}}^m p_{i'}\bigE{Z^{(\tc{i'})}_{i'}}
  \\
  & \stackrel{(c)}{=} &\sum_{i=1}^m\frac{p_{i}^2}{n^2_i}\biggl(\sum_{j=1}^{n_{i}}\bigE{Z^{(\tagg)}_{i,j}Z^{(\tc{i})}_{i,j}}+\sum_{j=1}^{n_i}\bigE{Z^{(\tagg)}_{i,j}}\sum_{j'=1}^{n_i}\bigE{Z^{(\tc{i})}_{i,j'}}\biggr) \nonumber\\
  && +\>\sum_{i=1}^m p_i\bigE{Z^{(\tagg)}_{i}}\sum_{\substack{i'\neq i}}^m p_{i'}\bigE{Z^{(\tc{i'})}_{i'}}
  \\
  & = &\sum_{i=1}^m\frac{p_{i}^2}{n^2_i}\biggl(n_{i}\bigE{Z^{(\tagg)}_{i}Z^{(\tc{i})}_{i}}+\sum_{j=1}^{n_i}\bigE{Z^{(\tagg)}_{i,j}}\sum_{j'\neq j}^{n_i}\bigE{Z^{(\tc{i})}_{i,j'}}\biggr) \nonumber\\
  && +\>\sum_{i=1}^m p_i\bigE{Z^{(\tagg)}_{i}}\sum_{\substack{i'\neq i}}^m p_{i'}\bigE{Z^{(\tc{i'})}_{i'}}
  \\
  & = &\sum_{i=1}^m\frac{p_{i}^2}{n^2_i}\Bigl(n_{i}\bigE{Z^{(\tagg)}_{i}Z^{(\tc{i})}_{i}}+n_i(n_i-1)\bigE{Z^{(\tagg)}_{i}}\bigE{Z^{(\tc{i})}_{i}}\Bigr) \nonumber\\
  && +\>\sum_{i=1}^m p_i\bigE{Z^{(\tagg)}_{i}}\sum_{\substack{i'\neq i}}^m p_{i'}\bigE{Z^{(\tc{i'})}_{i'}},
  \IEEEyesnumber\label{eq:second-term_practical-weight}
\end{IEEEeqnarray*}
where $(a)$ holds since the $\E{L_i}=0$ and $W^{(\tagg)}\indep L_i$, for all $i\in [m]$, $(b)$ is due to the fact that $W_{i}^{(\tagg)}\indep\vect{W}_{i'}^{(\tc{i'})}$ whenever $i'\neq i$, and $(c)$ satisfies as $Z_{i,j}^{(\tagg)}\indep\vect{Z}_{i,j'}^{(\tc{i})}$ for $j'\neq j$.

Note that in $(c)$, we have
\begin{IEEEeqnarray*}{rCl}
  \bigE{Z^{(\tagg)}_{i,j}Z^{(\tc{i})}_{i,j}}& = &\sum_{z_i\in\set{Z}_i} z_i\BigEcond{Z^{(\tagg)}_{i,j}}{Z^{(\tc{i})}_{i,j}=z_i}\BigPrv{Z^{(\tc{i})}_{i,j}=z_i}
  \\
  & = &\sum_{z_i\in\set{Z}_i}z_i\sum_{z'_i\in\set{Z}_i}z'_i\BigPrvcond{Z^{(\tagg)}_{i,j}=z'_i}{Z^{(\tc{i})}_{i,j}=z_i}\BigPrv{Z^{(\tc{i})}_{i,j}=z_i}
  \\
  & = &\sum_{z_i\in\set{Z}_i}\sum_{z'_i\in\set{Z}_i}z_i z'_i\cdot P_{Z^{(t_i)}_i\mid Z^{(0)}_i}(z'_i\mid z_i) P_{Z^{(0)}_i}(z_i)
  \\
  & = &\sum_{z_i\in\set{Z}_i}\sum_{z'_i\in\set{Z}_i}z_i z'_i\cdot p_{z_i,z_i}^{t_i} P_{Z^{(0)}_i}(z_i)=\bigE{Z^{(\tagg)}_{i}Z^{(\tc{i})}_{i}},
\end{IEEEeqnarray*}
which is independent of $j\in [n_i]$.

Moreover, regarding the first term in~\eqref{eq:all-terms_practical-weight}, one can easily see that
\begin{IEEEeqnarray*}{rCl}
  \bigE{\tilde{W}^2}& = &\bigVar{\tilde{W}}+\bigl(\bigE{\tilde{W}}\bigr)^2
  \\
  & = &\sum_{i=1}^m p^2_i\Bigl(\biggVar{\frac{1}{n_i}\sum_{j=1}^{n_i} Z_{i,j}^{(\tc{i})}} + \eVar{L_i}\Bigr)+\biggl(\sum_{i=1}^m p_i\bigl(\bigE{Z_{i}^{(\tc{i})}}+\eE{L_i}\bigr)\biggr)^2
  \\
  & = &\sum_{i=1}^m p^2_i\Bigl(\biggVar{\frac{1}{n_i}\sum_{j=1}^{n_i} Z_{i,j}^{(\tc{i})}} + \eVar{L_i}\Bigr)+\biggl(\sum_{i=1}^m p_i\bigl(\bigE{Z_{i}^{(\tc{i})}}+\eE{L_i}\bigr)\biggr)^2
  \\
  & = &\sum_{i=1}^m p^2_i\Bigl(\frac{1}{n_i}\bigVar{Z_{i}^{(\tc{i})}} + \eVar{L_i}\Bigr)+\biggl(\sum_{i=1}^m p_i\bigl(\bigE{Z_{i}^{(\tc{i})}}+\underbrace{\eE{L_i}}_{=0}\bigr)\biggr)^2
  \\
  & = &\sum_{i=1}^m p^2_i\frac{\bigVar{Z_{i}^{(\tc{i})}}}{n_i}+\sum_{i=1}^m p^2_i\eVar{L_i}+\biggl(\sum_{i=1}^m p_i\bigE{Z_{i}^{(\tc{i})}}\biggr)^2
  \\
  & = &\sum_{i=1}^m p^2_i\frac{\bigVar{Z_{i}^{(\tc{i})}}}{n_i}+\sum_{i=1}^m p^2_i(2\eta_i^2)+\biggl(\sum_{i=1}^m p_i\bigE{Z_{i}^{(\tc{i})}}\biggr)^2
  \\
  & \stackrel{(d)}{=} &\sum_{i=1}^m p^2_i\frac{\bigVar{Z_{i}^{(\tc{i})}}}{n_i}+\sum_{i=1}^m 2 p^2_i\left(\frac{z_{i,\textnormal{max}}-z_{i,\textnormal{min}}}{n_i g_i(t_i,\bar{\eps})}\right)^2+\biggl(\sum_{i=1}^m p_i\bigE{Z_{i}^{(\tc{i})}}\biggr)^2,\IEEEyesnumber\label{eq:first-term_practical-weight}
\end{IEEEeqnarray*}
where $(d)$ follows from~\eqref{eq:variance_Laplace-RV}.

Hence, combining~\eqref{eq:second-term_practical-weight} and~\eqref{eq:first-term_practical-weight},~\eqref{eq:all-terms_practical-weight} becomes
\begin{IEEEeqnarray*}{rCl}
  \bigE{\hat{\Loss}(\tilde{\vect{W}}(\set{T}_\txtc,\set{E}_\txtc),\set{D}^{(\tagg)})}& = &
  \sum_{i=1}^m p^2_i\frac{\bigVar{Z_{i}^{(\tc{i})}}}{n_i}+\biggl(\sum_{i=1}^m p_i\bigE{Z_{i}^{(\tc{i})}}\biggr)^2+\sum_{i=1}^m 2 p^2_i\left(\frac{z_{i,\textnormal{max}}-z_{i,\textnormal{min}}}{n_i g_i(t_i,\bar{\eps})}\right)^2
  \nonumber\\
  && -\>2\sum_{i=1}^m\frac{p_{i}^2}{n^2_i}\Bigl(n_{i}\bigE{Z^{(\tagg)}_{i}Z^{(\tc{i})}_{i}}+n_i(n_i-1)\bigE{Z^{(\tagg)}_{i}}\bigE{Z^{(\tc{i})}_{i}}\Bigr)
  \nonumber\\
  && -\>2\sum_{i=1}^m p_i\bigE{Z^{(\tagg)}_{i}}\sum_{\substack{i'\neq i}}^m p_{i'}\bigE{Z^{(\tc{i'})}_{i'}}
  +\sum_{i=1}^m p_i\bigE{\bigl(Z^{(\tagg)}_{i}\bigr)^2}.\IEEEyesnumber\label{eq:final_all-terms_practical-weight}
\end{IEEEeqnarray*}

\smallskip

\noindent
\textbf{Loss for ideal weight:} Next, we consider the expected empirical loss for $W^{(\tagg)}$, which is given by
\begin{IEEEeqnarray*}{rCl}  
  \bigE{\hat{\Loss}(\vect{W}^{(\tagg)},\set{D}^{(\tagg)})}& = &\sum_{i=1}^m p_i\BigE{\hat{\Loss}_{i}(\vect{W}^{(\tagg)},\set{D}_i^{(\tagg)})}
  \\
  & = &\sum_{i=1}^m p_i\biggE{\frac{1}{n_i}\sum_{j=1}^{n_i}\Bigl(W^{(\tagg)}-Z_{i,j}^{(\tagg)}\Bigr)^2}
  \\
  & = &\sum_{i=1}^m p_i\frac{1}{n_i}\biggE{\sum_{j=1}^{n_i}\bigl(W^{(\tagg)}\bigr)^2-2 W^{(\tagg)}\sum_{j=1}^{n_i}Z_{i,j}^{(\tagg)}+\sum_{j=1}^{n_i}\bigl(Z_{i,j}^{(\tagg)}\bigr)^2}
  \\
  & = &\sum_{i=1}^m p_i\frac{1}{n_i}\biggl(\sum_{j=1}^{n_i}\BigE{\bigl(W^{(\tagg)}\bigr)^2}-2\BigE{W^{(\tagg)}\bigl(n_i W^{(\tagg)}_i\bigr)}+\sum_{j=1}^{n_i}\BigE{\bigl(Z_{i,j}^{(\tagg)}\bigr)^2}\biggr)
  \\
  & = &\sum_{i=1}^m p_i\frac{1}{n_i}\Bigl(n_i\bigE{\bigl(W^{(\tagg)}\bigr)^2}-2n_i\bigE{W^{(\tagg)}W_i^{(\tagg)}}+n_i\bigE{\bigl(Z_{i}^{(\tagg)}\bigr)^2}\Bigr)
  \\
  & = &\sum_{i=1}^m p_i\Bigl(\bigE{\bigl(W^{(\tagg)}\bigr)^2}-2\bigE{W^{(\tagg)}W_i^{(\tagg)}}+\bigE{\bigl(Z_{i}^{(\tagg)}\bigr)^2}\Bigr)
  \\
  & = &\bigE{\bigl(W^{(\tagg)}\bigr)^2}-2\biggE{W^{(\tagg)}\sum_{i=1}^m p_i W_i^{(\tagg)}}+\sum_{i=1}^m p_i\bigE{\bigl(Z_{i}^{(\tagg)}\bigr)^2}
  \\
  & = &-\bigE{\bigl(W^{(\tagg)}\bigr)^2}+\sum_{i=1}^m p_i\bigE{\bigl(Z_{i}^{(\tagg)}\bigr)^2}.
  \IEEEyesnumber\label{eq:all-terms_ideal-weight}
\end{IEEEeqnarray*}

Moreover, one can easily obtain
\begin{IEEEeqnarray*}{rCl}
  \bigE{\bigl(\vect{W}^{(\tagg)}\bigr)^2}& = &\bigVar{\vect{W}^{(\tagg)}}+\bigl(\bigE{\vect{W}^{(\tagg)}}\bigr)^2
  \\
  & = &\sum_{i=1}^m p_i^2\biggVar{\frac{1}{n_i}\sum_{j=1}^{n_i} Z_{i,j}^{(\tagg)}}+\biggl(\sum_{i=1}^m p_i\bigE{Z_{i}^{(\tagg)}}\biggr)^2
  \\
  & = &\sum_{i=1}^m p^2_i\frac{\bigVar{Z_i^{(\tagg)}}}{n_i}+\biggl(\sum_{i=1}^m p_i\bigE{Z_{i}^{(\tagg)}}\biggr)^2.\IEEEyesnumber\label{eq:first-term_ideal-weight}
\end{IEEEeqnarray*}

Therefore, combining~\eqref{eq:first-term_ideal-weight}, \eqref{eq:all-terms_ideal-weight} becomes
\begin{IEEEeqnarray*}{c}
  \bigE{\hat{\Loss}(\vect{W}^{(\tagg)},\set{D}^{(\tagg)})}
  =-\sum_{i=1}^m p^2_i\frac{\bigVar{Z_i^{(\tagg)}}}{n_i}-\biggl(\sum_{i=1}^m p_i\bigE{Z_{i}^{(\tagg)}}\biggr)^2
  +\sum_{i=1}^m p_i\bigE{\bigl(Z_{i}^{(\tagg)}\bigr)^2}.\IEEEyesnumber\label{eq:final_all-terms_ideal-weight}
\end{IEEEeqnarray*}
Finally, the expected loss difference between~\eqref{eq:final_all-terms_practical-weight} and~\eqref{eq:final_all-terms_ideal-weight} becomes
\begin{IEEEeqnarray*}{rCl}
  \IEEEeqnarraymulticol{3}{l}{%
    \bigE{\hat{\Loss}(\tilde{\vect{W}}(\set{T}_\txtc,\set{E}_\txtc),\set{D}^{(\tagg)})}-\bigE{\hat{\Loss}(\vect{W}^{(\tagg)},\set{D}^{(\tagg)})}}\nonumber\\*\quad%
  & = &\sum_{i=1}^m 2 p^2_i\left(\frac{z_{i,\textnormal{max}}-z_{i,\textnormal{min}}}{n_i g_i(t_i,\bar{\eps})}\right)^2
  +\sum_{i=1}^m p^2_i\frac{\bigVar{Z_{i}^{(\tc{i})}}}{n_i}+\biggl(\sum_{i=1}^m p_i\bigE{Z_{i}^{(\tc{i})}}\biggr)^2
  \nonumber\\
  && -\>2\sum_{i=1}^m\frac{p_{i}^2}{n^2_i}\Bigl(n_{i}\bigE{Z^{(\tagg)}_{i}Z^{(\tc{i})}_{i}}+n_i(n_i-1)\bigE{Z^{(\tagg)}_{i}}\bigE{Z^{(\tc{i})}_{i}}\Bigr)
  \nonumber\\
  && -\>2\sum_{i=1}^m p_i\bigE{Z^{(\tagg)}_{i}}\sum_{\substack{i'\neq i}}^m p_{i'}\bigE{Z^{(\tc{i'})}_{i'}}
  +\sum_{i=1}^m p^2_i\frac{\bigVar{Z_i^{(\tagg)}}}{n_i}+\biggl(\sum_{i=1}^m p_i\bigE{Z_{i}^{(\tagg)}}\biggr)^2.
  \IEEEeqnarraynumspace\label{eq:loss-difference_mean-estimation}
\end{IEEEeqnarray*}
\end{IEEEproof}
}}{}

\subsection{Proposed Federated Learning Protocol}
\label{subsec:Proposed_scheduling}
Given the privacy requirement $\bar{\epsilon}$ and the model deployment time $t_{\textnormal{agg}}$, our FL protocol is as follows:
\begin{itemize}
\item \textbf{Step 1 (Scheduling of data collection time):} The PS schedules the data collection time $\set{T}_\txtc$ and informs the client $i$ its scheduled time $t_{\textnormal{c},i}$.
\item  \textbf{Step 2 (Local training):} The client $i\in[m]$ collects data at its scheduled time $t_{\textnormal{c},i}$ and performs ERM to train the local model.
\item \textbf{Step 3 (Adaptive noise power calculation):} The client $i\in[m]$ computes its $\epsilon_{\textnormal{c},i}$ based on its $\Delta_i(t_{\textnormal{agg}}-t_{\textnormal{c},i})$
  \begin{equation}
    \epsilon_{\textnormal{c},i} = \ln \left( \frac{1}{\Delta_i(t_{\textnormal{agg}}-t_{\textnormal{c},i})} (\exp( \bar{\epsilon}) - 1) +1 \right),
    \label{eqn:def_epsilon_c}
  \end{equation}
  where~\eqref{eqn:def_epsilon_c} is obtained from~\eqref{eqn:expression_epsilon_t}. This $\epsilon_{\textnormal{c},i}$ represents the amount of DP that the classical mechanism needs to provide to maintain the overall DP requirement $\bar{\epsilon}$. In this paper, we adopt the Laplace mechanism, and add at client $i$ a Laplace noise with parameter $\eta_i\equiv\nicefrac{s_{i}^{(f)}}{\epsilon_{\textnormal{c},i}}$, which follows the distribution,
  \begin{equation*}
    p_{\tn{L}}(x;\eta_i) = \frac{1}{2\eta_i}\exp\left(-\frac{\eabs{x}}{\eta_i} \right),
  \end{equation*}
  where $s_i^{(f)}$ is the $\ell_1$-sensitivity of the underlying learning algorithm $f$ (we adopt ERM in Section~\ref{sec:discussion_simulation}) given by
  \begin{equation*}
    s_i^{(f)}\eqdef\max_{\mathcal{D}_i,\mathcal{D}_i'}\norm[1]{f(\mathcal{D}_i) - f(\mathcal{D}_i')},\quad i\in [m].
  \end{equation*}
\end{itemize}

With the above analysis, we can now use the upper bound in \eqref{eqn:UB_loss} as a surrogate to solve the optimization problem in \eqref{eqn:def_loss_gap}. Specifically, for each scheduling policy $\set{T}_\txtc$, we use \eqref{eqn:UB_loss} to calculate an upper bound on the loss difference. We repeat this for every policy $\set{T}_\txtc$ and find $\set{T}^\ast_\txtc$ that minimizes the upper bound. It is obvious that as the number of clients and how old the data we can tolerate increases, the computational complexity becomes unacceptable. Hence, it is crucial to design low-complexity algorithms that can efficiently find optimal scheduling, which is left for future work. Moreover, extending our analytic results to the conventional FL differentially-private gradient descent algorithms with more than one iteration is one of our future work.


\section{Discussion and Simulation}
\label{sec:discussion_simulation}

In this section, we present simulation results to validate our analysis and showcase the efficacy of the proposed FL protocol. We first introduce the simulation setting in Section~\ref{subsec:setting} and then discuss our results in Section \ref{subsec:results}.

\subsection{Setting}
\label{subsec:setting}

We consider a scenario similar to the electricity consumption forecasting experiment described in~\cite{ZhangWeiBerryHuang24_1}, where a PS aims to train a global model to estimate the average electricity consumption of households in London. Under a privacy-sensitive system, we assume three clients exist in the system. i.e., $m=3$. Unlike~\cite{ZhangWeiBerryHuang24_1}, we assume each client's electricity consumption at scheduled time $t_{\textnormal{c},i}$ follows $\pi_i$ and the transition probability matrix from $t_{\textnormal{c},i}+j-1$ to $t_{\textnormal{c},i}+j$ is $\mat{P}_i$ for every $j\in\mathbb{N}$, both may vary from client to client. Specifically, the initial probability distributions for the three clients are set to be {\hy $\mu_1^{(\tc{1})} = [0.8,0.2,0,0]$, $\mu^{(\tc{2})}_2 = [0,0.1,0.5,0.4]$ and $\mu^{(\tc{3})}_3 = [0.2,0.3,0.5,0]$}, respectively.

In our system, each client samples 100 data at time $t_{\textnormal{c},i}$ individually and performs local training using the ERM algorithm with the mean square error loss function to achieve optimal local weights. Subsequently, it uploads these weights to the PS. We assume that only one client can upload the local weights to the PS at a time. After receiving all clients’ weights, the PS aggregates them and obtains the global weights at time $\tagg$.

To simplify the problem, we quantize the electricity consumption per half hour into four possible values: 20, 50, 100, and 200 (in watt-hour) {\hy for all clients. Moreover, we assume that the variation in electricity consumption can be formulated as a four-state birth-and-death Markov chain, where the transition probability matrix for each client $i$ is expressed as:
\begin{IEEEeqnarray*}{c}
  \mat{P}_i =
  \begin{pmatrix}
    1-p_i-q_i & p_i+q_i & 0 & 0 \\
    q_i & 1-p_i-q_i & p_i & 0 \\
    0 & q_i & 1-p_i-q_i & p_i \\
    0 & 0 & p_i+q_i & 1-p_i-q_i \\
  \end{pmatrix},    
\end{IEEEeqnarray*}
where $\vect{p}=(p_1,p_2,p_3)=(0.1,0.2,0.4)$ and $\vect{q}=(q_1,q_2,q_3) = (0.2,0.1,0.4)$.} The client $i$ collects data at time $t_{\textnormal{c},i}$ and the initial state is then given.
For the maximal total variation $\Delta_i(t)$, $i\in [3]$, $t\in\Naturals$, we apply an upper bound in \cite[Prop.~5]{ZhangWeiBerryHuang24_1}. It is expressed as
\begin{equation*}
  \Delta_i(t)\leq\min\left\{1, \max_{x_i\in\set{Z}_i}\sqrt{\frac{1-\pi_{i}(x_i)}{\pi_{i}(x_i)}}(\gamma_{i,*})^t\right\},
\end{equation*}
{\hy where $1-\gamma_{i,*}$ represents the \emph{absolute spectral gap} of client $i$’s time-varying database, i.e., $\gamma_{i,*}\eqdef\max\{\gamma_{i,2}, \eabs{\gamma_{i,4}}\}$, $\gamma_{i,l}$ is the $l$-th large eigenvalue of the transition probability matrix $\mat{P}_i$, $l\in [4]$~\cite{LevinPeresWilmer17_1}.}

\subsection{Results}
\label{subsec:results}

{\hy
We present the simulation results in terms of the expected loss difference~\eqref{eqn:def_loss_gap} based on different scheduling schemes. The considered schemes are as follows.
\begin{enumerate}
\item[1)] Random scheduling $+$ constant noise: We randomly select a set of scheduled times $\set{T}=\{\tc{i}\}_{i\in [m]}$ and a constant noise parameter, $\epsc{i}=\eps_{\txtc}$, $\forall\,i\in [m]$, is used for every client regardless of $t_{\textnormal{c},i}$. The $\bar{\eps}$ is then chosen to be $\bar{\eps}=\max_{i\in [m]}\{\eps(\tagg-\tc{i},\eps_{\txtc})\}$ based on~\eqref{eqn:expression_epsilon_t}.

\item[2)] Random scheduling $+$ adaptive noise: We randomly select a set of scheduled times $\set{T}_{\txtc}$ and each client adds a noise according to~\eqref{eqn:def_epsilon_c}.
  
\item[3)] Proposed scheduling $+$ constant noise: Fix a constant noise level $\eps_{\txtc}$ for each client, find the best set of scheduled times $\set{T}_\txtc$ that achieves the minimum loss difference according to Theorem~\ref{thm:UB_convergence}, and use this policy to determine its achievable $\bar{\epsilon}=\max_{i\in [m]}\{\eps(\tagg-\tc{i},\eps_{\txtc})\}$.

\item[4)] Proposed scheduling $+$ adaptive noise: For each set of scheduled times $\set{T}_{\txtc}$, we adaptively add noise so that $\bar{\epsilon}$ is satisfied according to~\eqref{eqn:def_epsilon_c}. We then find the best scheduling that results in minimum loss difference according to Theorem~\ref{thm:UB_convergence}.

\item[5)] Optimal scheduling $+$ constant noise: Similar to 3) but use simulation for the expected loss difference according to~\eqref{eqn:def_loss_gap}  instead of Theorem~\ref{thm:UB_convergence}.

\item[6)] Optimal scheduling $+$ adaptive noise: Similar to 4) but use simulation for the expected loss difference according to~\eqref{eqn:def_loss_gap} instead of Theorem~\ref{thm:UB_convergence}. 
\end{enumerate}
}

\ifthenelse{\boolean{arXiv}}{
\begin{figure*}[t!]
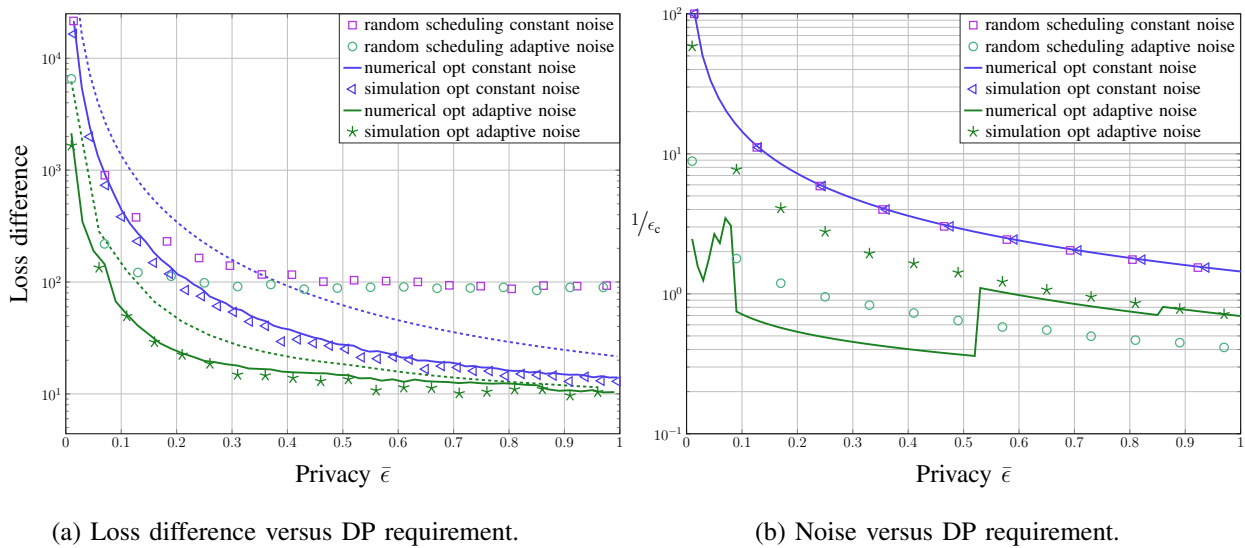

  \centering
  \begin{subfigure}{0.45\linewidth}
    \centering
    \input{\Figs/Fig_loss_privacy_new.tex}
    \subcaption{ Loss difference versus DP requirement.}
    \label{fig:acc_vs_privacy}
  \end{subfigure}
  \hfill
  \begin{subfigure}{0.5\linewidth}
    \input{\Figs/Fig_noise_privacy_new.tex} 
    \subcaption{ Noise versus DP requirement.}
    \label{fig:noise_vs_privacy}
  \end{subfigure}
  \caption{ Loss difference/Noise versus DP requirement.}
\end{figure*}
}{
\begin{figure}[t!]
  \centering
  \input{\Figs/Fig_loss_privacy_new.tex}
  \caption{Loss difference versus DP requirement.}
  \label{fig:acc_vs_privacy}
\end{figure}}

\ifthenelse{\boolean{NOTE}}{
\begin{figure}[t!]
  \centering
  \input{\Figs/bestScheduling_loss_difference_AgeDP_4MarkovChains_tagg10_v1.tex}
  \caption{\hy Loss difference versus DP requirement.}
  \label{fig:acc_vs_privacy_exact}
  \todo[inline]{Complete other curves in \ref{fig:acc_vs_privacy} based on the new simulation parameters in Sec.~\ref{subsec:setting}.}
\end{figure}
}{}

One can expect that 1) and 2) would not perform well as they do not take scheduling into their design. Moreover, 3) would not perform well even though scheduling is taken into account, because each client needs to add constant noise to meet the privacy requirement for the worst client, compromising the accuracy. On the other hand, the proposed approach in 4) will perform the best {\hy in terms of the upper bound on loss difference stated in Theorem \ref{thm:UB_convergence}, as it jointly considers scheduling and adaptive noise. Furthermore, 5) and 6) are simulated to showcase the effectiveness of Theorem~\ref{thm:UB_convergence}.} \ifthenelse{\boolean{arXiv}}{}{More details about how Theorem 1 is applied in our simulation can be found in~\cite{LinLinHsuHuang24_1sub}.}

In the simulation results depicted in Fig.~\ref{fig:acc_vs_privacy}, as anticipated, it is evident that approaches 1) and 2) exhibit subpar performance due to the absence of scheduling considerations in their design. Likewise, approach 3) does not excel even with scheduling incorporated, as each client introduces a constant noise to meet the privacy requirement for the worst client, consequently compromising accuracy. In contrast, the proposed approach in 4) outperforms all others. By simultaneously considering scheduling and adaptive noise, it achieves the lowest loss difference. This underscores the effectiveness of our comprehensive approach in striking a balance between privacy and accuracy. Moreover, comparing the simulation results of 3) with 5) and those of 4) with 6) reveals that Theorem~\ref{thm:UB_convergence} accurately captures the trade-off and guides us towards proper scheduling designs. This validation highlights the robustness and applicability of the theoretical framework proposed in the paper. \ifthenelse{\boolean{arXiv}}{
The simulation results in Fig.~\ref{fig:noise_vs_privacy} show the noise power that we have to add in each scheme. {\hy Here, $\eps_{\txtc}\eqdef\frac{1}{m}\sum_{i=1}^m\eps_{\txtc,i}$.} The noise power of the schemes that add a constant noise to meet the privacy requirement is always larger than adaptive noise schemes, whether the scheduling is taken into consideration or not.
}{}


\section{Conclusion}
\label{sec:conclusion}

This paper investigated the intricate relationship between data aging, DP, and accuracy within FL systems, particularly in the context of time-varying databases. We introduced an age-dependent upper bound on the FL loss difference, providing a theoretical foundation for our research. Leveraging this bound, we devised a scheduling algorithm to minimize the loss difference while adhering to age-dependent DP constraints. Our findings indicate that the proposed FL system surpasses FL systems that overlook aging into the design by concurrently optimizing both the data collection timing and the noise injection level. An interesting and important future work is to design low-complexity age-aware scheduling policies.



\ifthenelse{\boolean{arXiv}}{\balance}{
\IEEEtriggeratref{9}
}

\bibliographystyle{IEEEtran}
\bibliography{defshort1.bib,biblioHY.bib}


\ifthenelse{\boolean{arXiv}}{
\onecolumn
\appendices

\section{Proof of Theorem~\ref{thm:UB_convergence}}
\label{apx:UB_convergence}
{\hy
By the definition of the expected (w.r.t. to practical weight, noise, and ideal weight) accuracy loss difference between the global practical weight $\tilde{\vect{W}}(\set{T}_\txtc,\set{E}_\txtc)=\frac{1}{m}\sum_{i=1}^{m}\bigl(\vect{W}_i^{(\tc{i})}+N_i\bigr)=\vect{W}(\set{T}_\txtc,\set{E}_\txtc)+N$ under a certain privacy guarantee, and the ideal weight $\vect{W}^{(\tagg)}$, we have
\begin{IEEEeqnarray}{rCl}
  \IEEEeqnarraymulticol{3}{l}{%
    \bigE[\tilde{\vect{W}},\vect{W}^{(\tagg)},N]{\hat{\Loss}(\tilde{\vect{W}},\set{D}^{(\tagg)})-\hat{\Loss}(\vect{W}^{(\tagg)}, \set{D}^{(t_{\textnormal{agg}})})}}  
  \nonumber\\*\quad%
  & = &\bigE{\hat{\Loss}(\tilde{\vect{W}},\set{D}^{(\tagg)})-\avgLoss(\tilde{\vect{W}})+\avgLoss(\vect{W}^{(\tagg)})-\hat{\Loss}\bigl(\vect{W}^{(\tagg)},\set{D}^{(t_{\textnormal{agg}})})+\avgLoss(\tilde{\vect{W}}\bigr)-\avgLoss(\vect{W}^{(\tagg)})}
  \nonumber\\
  & = &\bigE{\hat{\Loss}(\tilde{\vect{W}},\set{D}^{(\tagg)})-\avgLoss(\tilde{\vect{W}})}
  +\bigE{\avgLoss(\vect{W}^{(\tagg)})-\hat{\Loss}(\vect{W}^{(\tagg)},\set{D}^{(t_{\textnormal{agg}})})}
  +\bigE{\avgLoss(\tilde{\vect{W}})-\avgLoss(\vect{W}^{(\tagg)})}
  \nonumber\\
  & = &-\bigE{\textnormal{gen}(\tilde{\vect{W}},\set{D}^{(t_{\textnormal{agg}})})}
  +\bigE{\textnormal{gen}(\vect{W}^{(\tagg)},\set{D}^{(t_{\textnormal{agg}})})}
  +\bigE{\Risk\bigl(\tilde{\vect{W}},\vect{W}^{(\tagg)}\bigr)}
  \label{eqn:pf_thm1_three_terms}
\end{IEEEeqnarray}
where $\textnormal{gen}(W,\set{D})\eqdef\avgLoss(\vect{W})-\hat{\Loss}(\vect{W},\set{D})$ and $\Risk(\vect{W})\eqdef\avgLoss(\vect{W})-\avgLoss(\vect{W}')$ represent the generalization error and excess risk between weights $\vect{W}$ and $\vect{W}'$, respectively.

To upper bound the first term of the right-hand side (RHS) in~\eqref{eqn:pf_thm1_three_terms}, we note that
\begin{IEEEeqnarray}{rCl}
  -\bigE{\textnormal{gen}(\tilde{\vect{W}},\set{D}^{(t_{\textnormal{agg}})})}
  & = &\bigE{\hat{\Loss}\bigl(W+N,\set{D}^{(\tagg)}\bigr)-\avgLoss(W+N)}
  \nonumber\\
  & \stackrel{(a)}{=} &\BiggE{\sum_{i=1}^m p_i\frac{1}{\bigcard{\set{D}_i^{(\tagg)}}}\sum_{j=1}^{\ecard{\set{D}_i^{(\tagg)}}}\ell\bigl(W+N, Z_{i,j}^{(\tagg)}\bigr)-\avgLoss(W+N)} 
  \nonumber\\
  & \stackrel{(b)}{=} &\BiggE{\sum_{i=1}^m p_i\frac{1}{\bigcard{\set{D}_i}}\sum_{j=1}^{\ecard{\set{D}_i}}\left(\ell\bigl(W+N, Z_{i,j}^{(t_{\textnormal{agg}})}\bigr)-\Loss_{\mu_i}(W+N)\right)} 
  \nonumber \\
  & = &\sum_{i=1}^m p_i\frac{1}{\bigcard{\set{D}_i}}\sum_{j=1}^{\ecard{\set{D}_i}}\BigE{\bigl(\ell(W+N, Z_{i,j}^{(\tagg)})-\Loss_{\mu_i}(W+N)\bigr)},\label{eq:1st_term_gen-bound}
\end{IEEEeqnarray}
where $(a)$ uses the definition of the empirical loss function and $(b)$ follows as $\ecard{\set{D}_i}=\bigcard{\set{D}_i^{(\tc{i})}}=\bigcard{\set{D}_{i}^{(\tagg)}}$ and $p_i=\nicefrac{\ecard{\set{D}^{(\tc{i})}}}{\card{\set{D}}}$. Here, $Z_{i,j}^{(\tagg)}\sim\mu_i$ for all $j\in[\ecard{\set{D}_i}]$, $i\in [m]$.

Next, we adopt a variational presentation of the KL divergence between two distributions $P$ and $Q$ over the same alphabet $\set{X}$ as follows (the so-called Donsker-Varadhan representation, see, e.g., \cite[Th.~1]{Belghazi-etal18_1} or~\cite[Cor.~4.15]{BoucheronLugosiMassart13_1}),
\begin{IEEEeqnarray}{c}
  \erelDf{P}{Q} = \sup_{f}\bigl\{\E[P]{f(X)}-\log\eE[Q]{\ope^{f(X)}}\bigr\},
  \label{eq:KL_Donsker-Varadhan-version}
\end{IEEEeqnarray}
where the supremum is taken over all functions such that $\eE[Q]{\ope^{f(X)}}$ exists. Let us consider the corresponding $P$ and $Q$ in~\eqref{eq:KL_Donsker-Varadhan-version} as $P_i=P_{W,N,Z_i^{(\tagg)}}$ and $Q_i=P_{W}\otimes P_{N} \otimes\mu_i$, respectively, and further choose $f_i=\lambda\bigl(\ell(W+N,Z_{i,j}^{(\tagg)})-\bigE{\ell(W+N,Z_{i,j}^{(\tagg)})}\bigr)$ for some value $\lambda$, $i\in [m]$. Then, applying~\eqref{eq:KL_Donsker-Varadhan-version} we can obtain
\begin{IEEEeqnarray}{rCl}
  \relDf{P}{Q}& \geq &\eE[P]{f_i(X)}-\log{\bigE[Q]{\ope^{f_i(X)}}}
  \nonumber\\
  & \stackrel{(c)}{\geq} &\eE[P]{f_i(X)}-\psi_i(\lambda),\label{eq:apply_KL-divergence-upper-bound}
\end{IEEEeqnarray}
where $(c)$ holds since by assumption, the cumulant generating function of $f_i$ over $Q_i$, $\log{\E[Q]{\ope^{f_i(X)}}}$, is upper bounded by $\psi_i(\lambda)$. Hence, \eqref{eq:apply_KL-divergence-upper-bound} further gives that
\begin{IEEEeqnarray}{rCl}
  \bigE{\lambda\bigl(\ell(W+N,Z_{i,j}^{(\tagg)})-\bigE{\ell(W+N,Z_{i,j}^{(\tagg)})}\bigr)}& \leq &\bigrelDf{P_{W,N,Z_{i}^{(\tagg)}}}{P_{W}\otimes P_{N}\otimes\mu_i} + \psi_i(\lambda).
  \IEEEeqnarraynumspace\label{eqn:UB_loss_by_KL}
\end{IEEEeqnarray}
Observer that the above KL divergence can be further expanded to
\begin{IEEEeqnarray}{rCl}
  \bigrelDf{P_{W,N,Z^{(\tagg)}_i}}{P_{W} \otimes P_N \otimes \mu_i}
  & = &\sum_{w,n,z} P_{W,N,Z^{(\tagg)}_i}(w,n,z)\log{\Biggl(\frac{P_{W,N,Z^{(\tagg)}_i}(w,n,z)}{P_W(w)P_N(n)\mu_i}\Biggr)}
  \nonumber\\
  & \stackrel{(d)}{=} &\sum_{w,n,z}P_{W,Z^{(\tagg)}_i}(w,z)P_{N}(n)\log{\Biggl(\frac{P_{W,Z^{(\tagg)}_i}(w,z)P_N(n)}{P_W(w)P_N(n)\mu_i}\Biggr)}
  \nonumber\\
  & = &\sum_{w,z} P_{W,Z^{(\tagg)}_i}(w,z)\log{\Biggl(\frac{P_{W,Z^{(\tagg)}_i}(w,z)}{P_W(w)P_{Z_{i,j}^{(\tagg)}}(z)}\frac{P_{Z_{i,j}^{(\tagg)}}(z)}{\mu}\Biggr)}
  \nonumber\\
  & = &\bigMI{W}{Z_{i,j}^{(\tagg)}} + \sum_{z}P_{Z_{i,j}^{(\tagg)}}(z)\log{\Biggl(\frac{P_{Z_{i,j}^{(\tagg)}}(z)}{\mu_i}\Biggr)}
  \nonumber\\
  & \stackrel{(e)}{=} &\bigMI{W}{Z_{i,j}^{(\tagg)}},\label{eqn:KL_divergence_derive}
\end{IEEEeqnarray}
where $(d)$ follows because $N$ is independent of $\vect{W}$ and $Z_i^{(\tagg)}$, and $(e)$ holds since $Z_{i,j}^{(\tagg)}\sim\mu_i$, $i\in [m]$.

Using~\eqref{eqn:KL_divergence_derive}, \eqref{eqn:UB_loss_by_KL} becomes
\begin{IEEEeqnarray*}{c}
  \bigE{\lambda\bigl(\ell(W+N,Z_{i,j}^{(\tagg)})-\bigE{\ell(W+N,Z_{i,j}^{(\tagg)})}\bigr)}
  \leq\bigMI{W}{Z_{i,j}^{(t_{\textnormal{agg}})}} + \psi_i(\lambda),
\end{IEEEeqnarray*}
and for some $\lambda\in [0,b_{+}]$, we have
\begin{IEEEeqnarray}{c}
  \bigE{\ell(W+N,Z_{i,j}^{(\tagg)})-\bigE{\ell(W+N,Z_{i,j}^{(\tagg)})}}\leq\frac{1}{\lambda}\bigl[\bigMI{W}{Z_{i,j}^{(t_{\textnormal{agg}})}}+\psi_i(\lambda)\bigr].\label{eqn:1st_term_bound_lambda}
\end{IEEEeqnarray}
Note that \eqref{eqn:1st_term_bound_lambda} can be further tightened by choosing the best value of $\lambda$, which gives
\begin{IEEEeqnarray}{rCl}
  \bigE{\ell(W+N,Z_{i,j}^{(\tagg)})-\bigE{\ell(W+N,Z_{i,j}^{(\tagg)})}}& \leq &\min_{\lambda\in [0,b_{+}]}\frac{1}{\lambda}\bigl[\bigMI{W}{Z_{i,j}^{(t_{\textnormal{agg}})}}+\psi_i(\lambda)\bigr]
  \nonumber\\
  & = &\psi_{i,+}^{*-1}\Bigl(\bigMI{W}{Z_{i,j}^{(t_{\textnormal{agg}})}}\Bigr).\label{eqn:1st_term_bound}
\end{IEEEeqnarray}
  
Therefore, combining \eqref{eq:1st_term_gen-bound} and \eqref{eqn:1st_term_bound}, the first term of the RHS in~\eqref{eqn:pf_thm1_three_terms} becomes
\begin{IEEEeqnarray}{c}
  -\bigE{\textnormal{gen}(\tilde{\vect{W}},\set{D}^{(t_{\textnormal{agg}})})}\leq\sum_{i=1}^m p_i \frac{1}{\bigcard{\set{D}_i}}\sum_{j=1}^{\ecard{\set{D}_i}}\psi^{*-1}_{i,+}\bigl(\bigMI{\vect{W}}{Z_{i,j}^{(t_\textnormal{agg})}}\bigr),
  \label{eq:first-term_loss-difference}
\end{IEEEeqnarray}
Using a similar argument as above, one can also obtain
\begin{IEEEeqnarray}{c}
  \bigE{\textnormal{gen}(\vect{W}^{(\tagg)},\set{D}^{(t_{\textnormal{agg}})})}\leq\sum_{i=1}^m p_i \frac{1}{\bigcard{\set{D}_i}}\sum_{j=1}^{\ecard{\set{D}_i}}\phi^{*-1}_{i,-}\bigl(\bigMI{\vect{W}^{(\tagg)}}{Z_{i,j}^{(\tagg)}}\bigr).
  \label{eq:second-term_loss-difference}
\end{IEEEeqnarray}
Thus, \eqref{eqn:pf_thm1_three_terms} becomes~\eqref{eqn:UB_loss} by combining~\eqref{eq:first-term_loss-difference} and~\eqref{eq:second-term_loss-difference}, which completes the proof.
}}{}

\end{document}